\documentclass[preprint,review,12pt]{elsarticle}

\usepackage{amssymb}
\usepackage{amsmath}
\usepackage{amsthm}
\usepackage{amsfonts}
\usepackage{array}
\usepackage[caption=false,font=normalsize,labelfont=sf,textfont=sf]{subfig}
\usepackage{textcomp}
\usepackage{stfloats}
\usepackage{url}
\usepackage{verbatim}
\usepackage{graphicx}
\usepackage{makecell}
\usepackage{multirow}
\usepackage{placeins}
\usepackage[table]{xcolor}
\usepackage{makecell}
\usepackage{pifont}
\usepackage{xcolor}
\usepackage{soul}
\usepackage{booktabs}
\usepackage{adjustbox}
\usepackage{geometry}
\usepackage{tabularx}
\usepackage{makecell}
\usepackage{multicol}
\usepackage{bm}
\usepackage[utf8]{inputenc}
\usepackage[T1]{fontenc}
\usepackage{hyperref}
\usepackage{nicefrac}
\usepackage{microtype}
\usepackage{amstext}
\usepackage{doi}

\journal{Computer Vision and Image Understanding}%

\begin{document}

\begin{frontmatter}

\title{Vision-Based Natural Language Scene Understanding for Autonomous Driving: An Extended Dataset and a New Model for Traffic Scene Description Generation}%

\author[label1]{Danial {Sadrian Zadeh}\corref{cor1}}%
\ead{dsadrian@uwaterloo.ca}%
\author[label1]{Otman A. {Basir}}%
\ead{obasir@uwaterloo.ca}%
\author[label1,label2]{Behzad {Moshiri}}%
\ead{moshiri@ut.ac.ir}%

\affiliation[label1]{
    organization={Department of Electrical and Computer Engineering, University of Waterloo},
    addressline={200 University Avenue West},
    city={Waterloo},
    postcode={N2L 3G1},
    state={Ontario},
    country={Canada}
}%

\affiliation[label2]{
    organization={School of Electrical and Computer Engineering, College of Engineering, University of Tehran},
    addressline={North Kargar Street},
    city={Tehran},
    postcode={1439957131},
    state={Tehran},
    country={Iran}
}%

\cortext[cor1]{Corresponding author}%

\begin{abstract}
    Traffic scene understanding is essential for enabling autonomous vehicles to accurately perceive and interpret their environment, thereby ensuring safe navigation. This paper presents a novel framework that transforms a single frontal-view camera image into a concise natural language description, effectively capturing spatial layouts, semantic relationships, and driving-relevant cues. The proposed model leverages a hybrid attention mechanism to enhance spatial and semantic feature extraction and integrates these features to generate contextually rich and detailed scene descriptions. To address the limited availability of specialized datasets in this domain, a new dataset derived from the BDD100K dataset has been developed, with comprehensive guidelines provided for its construction. Furthermore, the study offers an in-depth discussion of relevant evaluation metrics, identifying the most appropriate measures for this task. Extensive quantitative evaluations using metrics such as CIDEr and SPICE, complemented by human judgment assessments, demonstrate that the proposed model achieves strong performance and effectively fulfills its intended objectives on the newly developed dataset.
\end{abstract}%


\begin{highlights}
    \item Accurate traffic scene understanding is fundamental to safe navigation.%
    \item Generating natural language descriptions facilitates traffic scene understanding.%
    \item CIDEr and SPICE are the most suitable metrics for evaluating scene descriptions.%
    \item Combining MViTv2-S and xLSTM produces contextually meaningful descriptions.%
    \item A dedicated dataset, derived from BDD100K dataset, is developed for model training.%
\end{highlights}%


\end{frontmatter}


\section{Introduction}%
\label{sec:s01}%
Autonomous driving systems (ADSs) represent a significant advancement in the automotive sector, unlocking new opportunities for efficient and comfortable transportation. They require a comprehensive understanding of their surrounding environment, including both static and dynamic entities \cite{2023_Zhang_Survey}. This underscores the irrefutable importance of situational awareness in autonomous driving (AD), as it directly influences safe navigation and logical decision-making, thereby promoting overall road safety.%

The rapid development of smart vehicles has created a pressing need to interpret complex traffic environments to ensure safety and efficiency. Traditional ADSs primarily focus on detecting, segmenting, and classifying specific traffic entities \cite{2024_Yue_WGS-YOLO,2025_Huang_Light-YOLO,2025_Erabati_RetSeg3D}---such as pedestrians, vehicles, and traffic signs---but often overlook the necessity for a more holistic semantic understanding of traffic scenes. Recent advancements at the intersection of natural language processing (NLP) and computer vision have improved decision-making systems for safer navigation in autonomous vehicles (AVs) \cite{2024_Osman,2025_Abdulgalil}. This progress is exemplified by the adoption of image captioning methods, which provide richer semantic information about the traffic scene \cite{2021_Li,2023_Malla_DRAMA,2024_Zhang_TSIC-CLIP}.%

The primary motivation for equipping vehicles with tools capable of identifying and translating potential driving risks and traffic rules into natural language descriptions is to enable and enhance situational awareness in traffic scene understanding. Such descriptions ultimately facilitate safe navigation and logical decision-making by providing a high degree of transparency in AD. However, current approaches and the explanations or descriptions they generate are often too generic and insufficiently detailed to capture the nuanced relationships and contextual dependencies present in complex traffic environments. Therefore, there is a crucial demand for approaches that can generate accurate, context-aware, and semantically rich descriptions of the traffic scene.%

Another motivation for the present study is that relying solely on visual data reduces the computational overhead associated with processing large volumes of data from additional sensors, such as LiDAR or radar. Achieving effective traffic scene understanding based on visual data necessitates the creation and development of a domain-specific dataset grounded in human perception of the driving environment---a resource that is currently lacking in the field. Existing datasets are either not publicly available or are insufficiently representative and inclusive for such a critical application, often failing to comprehensively identify or isolate risk factors, or being designed for different purposes.%

It is also noteworthy that the generated descriptions and information can be disseminated to other vehicles to facilitate safe, cooperative driving---for example, by notifying other vehicles of traffic accidents occurring farther along the roadway \cite{2023_Mohammed}. Additionally, this information can be further converted into audio output to assist visually impaired individuals \cite{2019_Keyes}, thereby providing them with critical awareness of the driving environment. Such technology can also serve as a valuable utility for enhancing mobility for individuals with various disabilities.%

Challenges related to secure navigation and rational decision-making in AVs are multifaceted. One significant aspect revolves around perception and understanding of the environment, particularly under conditions of low visibility or poor weather, which can lead to misinterpretation or misdetection of entities and, consequently, to unsafe decisions or collisions. Another significant challenge arises from the unpredictability and dynamic nature of traffic environments, including rapidly changing traffic patterns or erratic behavior from other road users. This introduces additional complexity that can adversely affect navigation and decision-making processes in AVs. Furthermore, adhering to rules and regulations specific to different driving environments---such as understanding traffic signs and road markings---poses another challenge, especially when these indicators are faded or obscured, thus increasing the risk of hazardous navigation \cite{2023_Padmaja_Survey,2025_Shaoo_Review}. Lastly, managing large volumes of data from various sensors, including cameras, LiDAR, and radar, imposes significant computational demands and necessitates the use of sophisticated optimization algorithms to ensure computational efficiency.%

On the other hand, one of the major challenges in image captioning for traffic scene understanding is the lack of publicly available, captioned or annotated datasets, as many researchers are unwilling to share their data due to privacy or proprietary concerns. Moreover, examples of such unshared datasets, as discussed in related research, show that the reference captions often lack sufficient detail regarding critical aspects of the traffic environment, such as road conditions, traffic sign recognition, and potential risks. This limitation impedes the development of models capable of generating comprehensive and contextually rich descriptions. Additionally, not all available evaluation metrics for assessing the quality of generated descriptions---such as BLEU \cite{2002_Papineni_BLEU}, METEOR \cite{2004_Lavie_First_METEOR,2005_Banerjee_Second_METEOR,2007_Lavie_Third_METEOR}, and ROUGE \cite{2004_Lin_ROUGE}---are fully applicable to the area of traffic scene understanding, which involves a variety of dangers and risks. Compounding these issues, most caption generation models produce descriptions that are too brief to encompass all relevant details of a dynamic traffic environment. Addressing these gaps requires the development of detailed, annotated datasets tailored to traffic scenes, the design of domain-specific evaluation metrics, and advancements in models capable of generating longer and more informative contextual descriptions. These efforts are essential for improving situational awareness, navigation, and decision-making in AD.%

Given the motivations driving the development of AVs and the challenges associated with navigation and decision-making, it is necessary to develop a tool that provides descriptions of the driving environment from the driver's perspective to support traffic scene understanding. This involves interpreting the spatial and semantic relationships among entities in the traffic scene. The proposed tool would represent the driving scene through natural language descriptions, utilizing only visual input data. In this context, the concrete and measurable objectives pursued in the present study are as follows:%
\begin{itemize}
    \item \textbf{Model Development:} Design and implementation of an encoder-decoder architecture for generating natural language descriptions of static traffic scenes. The model is selected from ten combinations of five encoders and two decoders, incorporating a hybrid attention fusion mechanism to enhance the final output. The encoder is chosen from a pool of two convolutional neural network (CNN)-based models and three Transformer-based models, while the decoder is selected from two recurrent neural network (RNN) options: LSTM \cite{1997_Hochreiter_LSTM} and extended LSTM (xLSTM) \cite{2024_Beck_xLSTM}.%
    \item \textbf{Dataset Development:} Development of a specifically tailored dataset that encompasses all critical aspects of the driving scene. This is achieved by carefully designing guidelines that consider the key factors for ensuring a safe and reliable driving experience and by outsourcing the process of generating reference scene descriptions to individuals with driving experience to ensure contextual accuracy.%
    \item \textbf{Quantitative and Qualitative Evaluation:} Assessment of model performance using both standard metrics and human evaluation. This involves a comprehensive investigation and analysis of image captioning evaluation metrics to identify the most suitable metrics for description generation evaluation in the field of AD.%
\end{itemize}%

The remainder of the manuscript is organized as follows. Section~\ref{sec:s02} provides a comprehensive review of prior studies, while Section~\ref{sec:s03} outlines the proposed framework in detail, describing the architectural hypotheses, the dataset, and the experimental design to address the research objectives. Subsequently, Section~\ref{sec:s04} presents the results obtained, providing a critical interpretation of the findings through a comprehensive analysis and evaluation. Finally, Section~\ref{sec:s05} concludes the manuscript, summarizing the main contributions, discussing the implications, and providing a brief overview of the future work.

\section{Related Work}%
\label{sec:s02}%
This section presents a detailed review of prior studies as well as gaps in the literature related to traffic scene understanding through natural language text generation within the context of AD, which underpin the motivation for the current study.%

\subsection{Review of Prior Studies}%
\subsubsection{Natural Language Explanations for Vehicle Behavior}%
Kim et al. \cite{2018_Kim_BDD-X} proposed an end-to-end CNN-LSTM-Attention architecture to incorporate introspective explanations and provide natural language explanations describing the reasons for different behaviors of the ego vehicle. Essentially, a pre-trained controller, serving as a decision-making unit that controls the direction and acceleration of the vehicle, utilizes the visual attention mechanism to identify critical image regions. An LSTM-based explanation generator then generates natural language explanations that describe the reasons behind the behavior of the controller. Two major contributions of this work are the employment of attention alignment, which aligns the controller of the vehicle and the LSTM-based explanation generator by drawing their attention to the same regions in the input, and the creation of the BDD-X dataset \cite{2018_Kim_BDD-X}, which is built by collecting human-generated explanations for the BDDV dataset \cite{2017_Xu_BBDV}. The metrics used for caption evaluation are BLEU \cite{2002_Papineni_BLEU}, METEOR \cite{2004_Lavie_First_METEOR,2005_Banerjee_Second_METEOR,2007_Lavie_Third_METEOR}, and CIDEr \cite{2015_Vedantam_CIDEr}, along with human-based evaluation. One of the main limitations of this study is that the reference explanations are provided by a third-party observer about the ego vehicle, rather than the driver's viewpoint or perception of the scene.%

To increase the level of safety and passenger alertness for accident avoidance, generating captions for near-future events is particularly important. In light of this, Mori et al. \cite{2021_Mori} proposed an image captioning architecture that produces captions explaining near-future events based on images recorded from earlier to current moments. Additionally, the architecture takes as input the motion information of the ego vehicle, including velocity, acceleration, and steering angle, to monitor changes in the sequence of input images. This architecture first feeds the sequence of images to a CNN feature extractor, after which the extracted features, along with the motion information of the ego vehicle, are fed to an LSTM-based encoder. Based on the encoded information, an LSTM-based action regressor module predicts the near-future motion of the ego vehicle. Finally, an LSTM-based decoder outputs near-future-related captions based on the outputs of the encoder and action regressor module. The dataset employed for this experiment is the BDD-X dataset \cite{2018_Kim_BDD-X}, and the metrics used for caption evaluation are BLEU \cite{2002_Papineni_BLEU}, METEOR \cite{2004_Lavie_First_METEOR,2005_Banerjee_Second_METEOR,2007_Lavie_Third_METEOR}, and CIDEr \cite{2015_Vedantam_CIDEr}. Nevertheless, similar to \cite{2018_Kim_BDD-X}, the generated captions in \cite{2021_Mori} offer reasons for the behavior of the ego vehicle.%

\subsubsection{Risk Factor Identification and Driver Assistance}%
Concerned with a diverse range of risk factors during driving, Mori et al. \cite{2019_Mori_Attention_Neural_Baby_Talk} proposed an image captioning architecture as an assistant to the driver, aiming to avoid driving accidents by identifying risk factors and drawing the driver's attention to them. This architecture first detects the objects in the input image by a Faster R-CNN \cite{2017_Ren_Faster_R-CNN} pre-trained on the {MS COCO} dataset \cite{2014_Lin_MS_COCO}. The detected objects are then assigned a risk factor based on their categories, positions, and distances as indicated in the annotation rule set. Each image is automatically annotated with five ground-truth captions, with each caption focusing on a different risk factor. The generation of captions is based on the Neural Baby Talk \cite{2018_Lu_Neural_Baby_Talk} architecture, which incorporates an added attention mechanism. The authors curated their own dataset of 30,320 images, each with five automatically generated captions. The effectiveness of this approach is evaluated through a questionnaire as well as BLEU \cite{2002_Papineni_BLEU} and METEOR \cite{2004_Lavie_First_METEOR,2005_Banerjee_Second_METEOR,2007_Lavie_Third_METEOR}. One limitation of this study is that rule-based approaches, which focus on a limited number of predefined risk factors, may not capture all the nuances of risk assessment in complex driving scenarios.%

\subsubsection{Comprehensive Scene Captioning in ITS}%
Current traffic scene understanding methods primarily focus on detecting objects or signs, but they lack generality, strategic guidance, and multimodal integration for complex scenarios. To overcome these issues, Li et al. \cite{2021_Li} proposed an end-to-end architecture inspired by \cite{2017_Vinyals}, where a VGG-16-based encoder \cite{2014_Simonyan_VGG} extracts features from input images and an LSTM-based decoder generates captions based on the extracted features. Due to the absence of a publicly available traffic dataset for image captioning, the authors created their own dataset by writing one English sentence for each image in the LaRA TLR public dataset \cite{2012_Siogkas}, thus manually annotating 11,178 images. For the same reason and also to ensure fair performance evaluation of their architecture, they compared their architecture with the state-of-the-art on the Flickr30k dataset \cite{2014_Young_Flickr30k} and utilized BLEU \cite{2002_Papineni_BLEU} for caption evaluation. The authors believe that their work \cite{2021_Li} is the first work on image captioning in the field of intelligent transportation systems, and the generated captions appear to include the relative locations of traffic entities as well as potential dangers ahead. Nevertheless, a notable issue with this work is that only one caption is created per image, while popular image captioning datasets typically include at least five captions per image. Providing multiple captions per image when creating such datasets is beneficial as it potentially allows the architecture to focus on one part of the scene at a time, and it is also known that the automatic evaluation metrics exhibit a stronger correlation with human judgment when provided with an increased number of reference captions \cite{2015_Vedantam_CIDEr}.%

\subsubsection{Instance Segmentation-Based Captioning}%
Srihari \& Sikha \cite{2022_Srihari_Mask-RCNN-UNET} proposed an instance segmentation-based image captioning architecture to generate captions for the Cityscapes dataset \cite{2016_Cordts_Cityscapes}. The dataset does not include ground-truth captions, and the authors did not provide any human-generated reference captions. In this architecture, the output captions are generated without sequence modeling, relying solely on the information provided by the instance segmentation model (i.e., color, distance, and location of objects), which is technically a Mask R-CNN combined with U-Net. Caption generation, in this work, mainly occurs through three modules. The first module calculates the distance between objects or scene entities; the second module determines the color of objects; and the third module identifies the location of objects by dividing the input image into five regions. This work does not report the employment of any metrics for caption evaluation. Additionally, the fixed frame structure of the captions may limit the flexibility and naturalness of the generated sentences.%

\subsubsection{Multimodal and Advanced Architectures}%
To evaluate the effectiveness of their proposed dataset, DRAMA, the authors of \cite{2023_Malla_DRAMA} proposed the DRAMA architecture. This architecture consists of an encoder that takes as input a frame of the scene and its corresponding optical flow; a ResNet-101 \cite{2016_He_ResNet} then processes these inputs, and the extracted features are concatenated and fed to the decoder, which comprises an LSTM with a soft attention mechanism for generating captions, as well as an MLP head for object localization. The effectiveness of this approach is evaluated through BLEU \cite{2002_Papineni_BLEU}, METEOR \cite{2004_Lavie_First_METEOR,2005_Banerjee_Second_METEOR,2007_Lavie_Third_METEOR}, ROUGE \cite{2004_Lin_ROUGE}, CIDEr \cite{2015_Vedantam_CIDEr}, and SPICE \cite{2016_Anderson_SPICE}. Nonetheless, the generated descriptions do not appear to fully include all aspects of the traffic scene.%

Describing scenes in traffic scenarios presents numerous challenges, including imprecise captions, unadapted descriptions, and a substantial number of model parameters. In light of these issues, Zhang et al. \cite{2024_Zhang_TSIC-CLIP} proposed Traffic Scene Image Captioning model based on Contrastive Language-Image Pretraining (TSIC-CLIP), which mainly consists of a feature extraction module and an image captioning module. In the first module, a CLIP-based \cite{2021_Radford_CLIP} adapter layer outputs global image features, a Faster R-CNN \cite{2017_Ren_Faster_R-CNN} detects the image objects, and a CLIP-based \cite{2021_Radford_CLIP} text encoder outputs text features taken from the text dataset. These three different outputs are then combined and fed to the second module for multimodal information fusion. The second module is essentially a Transformer \cite{2017_Vaswani_Transformer} with the attention layer replaced by a Global Weighted Attention Pooling (WGA-Pooling) layer, which comprises an encoder and decoder, to address the issue of a considerable number of model parameters. The authors evaluated the efficiency of their architecture through BLEU \cite{2002_Papineni_BLEU}, METEOR \cite{2004_Lavie_First_METEOR,2005_Banerjee_Second_METEOR,2007_Lavie_Third_METEOR}, ROUGE-L \cite{2004_Lin_ROUGE}, and CIDEr \cite{2015_Vedantam_CIDEr} on three datasets of Flickr30k \cite{2014_Young_Flickr30k}, {MS COCO} Captions \cite{2015_Chen_MS_COCO_Captions}, and BUUISE-Image \cite{2024_Zhang_TSIC-CLIP}.%

\subsection{Gaps in the Literature}%
Although prior studies have explored various aspects of traffic scene understanding through natural language text generation within the context of AD, several challenges remain to be addressed, as outlined below:%
\begin{itemize}
    \item There remains a significant need to enhance the extraction of task-specific features from the visual inputs and to generate longer, more detailed descriptions based on those inputs.%
    \item Dataset accessibility and performance generalizability have also posed significant challenges in prior studies. The annotated existing datasets are not only largely inaccessible to the public but also insufficiently detailed to capture all critical aspects of the scene. Furthermore, these datasets often lack diversity and reliability.%
    \item Most of the evaluation metrics for image captioning are not well-suited to traffic scene understanding. Prior studies have largely overlooked the importance of selecting appropriate evaluation metrics for this specific task, despite the necessity of relying on measures that accurately reflect task requirements. Therefore, reporting all available metrics for traffic scene understanding is unwarranted. There is a clear need to identify and employ metrics that are truly appropriate for this domain.%
\end{itemize}%

Consequently, the present study proposes a new encoder-decoder model with a special focus on enhancing the extraction of features from input images; it constructs a new dataset guided by the proposed annotation guidelines; and it investigates the appropriateness of various evaluation metrics to address the existing gaps in this domain.

\section{Methodology and Framework}%
\label{sec:s03}%
\subsection{Proposed Theoretical Framework}%
\begin{figure}[!t]
    \centering%
    \includegraphics[width=1.00\textwidth]{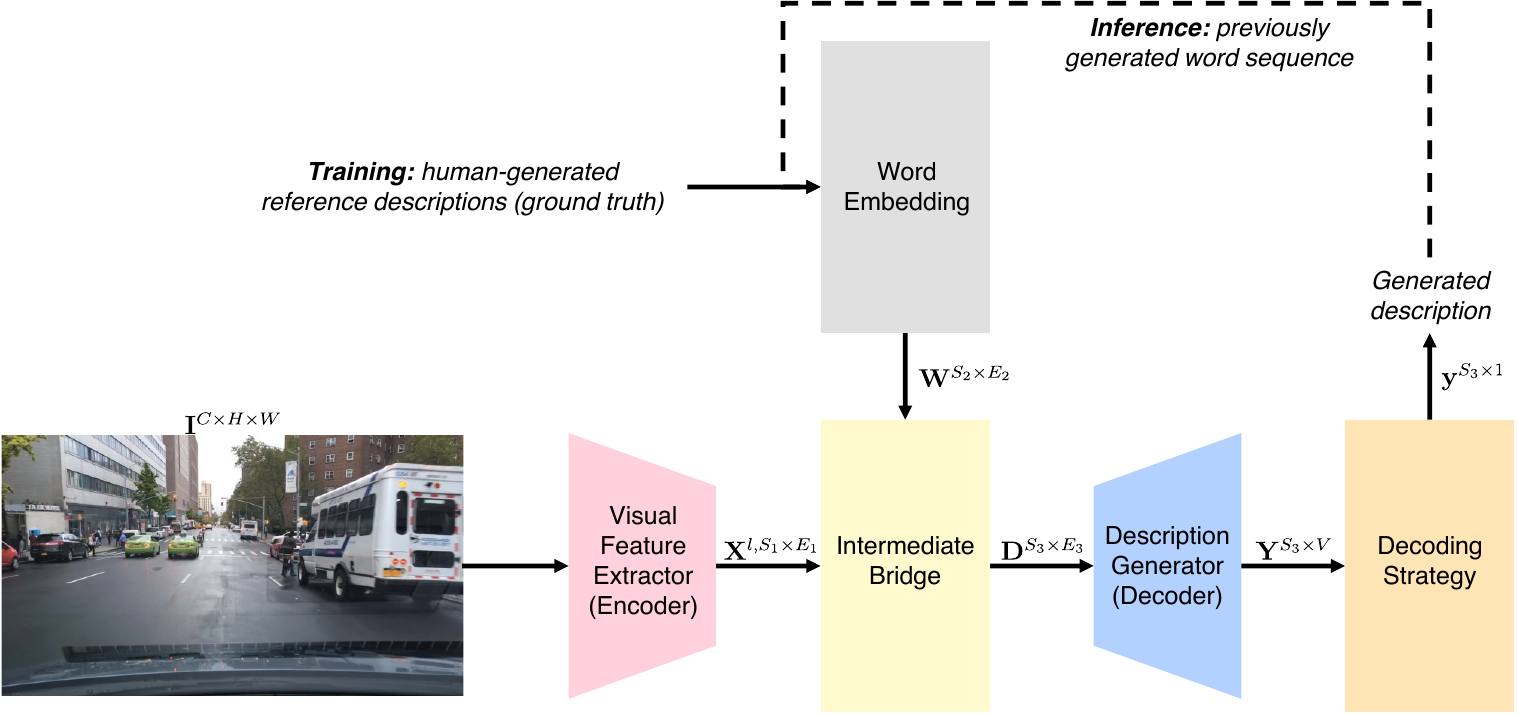}%
    \caption{Overall description generation architecture for implementing the proposed framework.}%
    \label{fig:encoder-decoder}%
\end{figure}

The task of describing the traffic scene from image inputs is fundamentally based on image captioning and can be formalized as the problem of learning a mapping function $f: \mathcal{I} \rightarrow \mathcal{Y}$, where $\mathcal{I}$ represents the image domain and $\mathcal{Y}$ represents the textual description domain. In the encoder-decoder paradigm, this mapping can be portrayed as:%
\begin{equation}\label{imgcap}
    P\left(\mathbf{y} | \mathbf{I}\right) = \prod\limits_{s=1}^{S} P\left(y_{s} | \mathbf{y}_{< s}, \mathbf{I}\right)\text{,}%
\end{equation}%
where $P\left(y_{s} | \mathbf{y}_{< s}, \mathbf{I}\right)$ stands for the probability of generating word (also called token) $y_{s}$ given the input image $\mathbf{I}$ and all previously generated words (tokens) $\mathbf{y}_{< s}$. The components of the framework are systematically introduced in the following subsubsections, with each step accompanied by a rationale explaining its significance. The overall architecture for implementing the proposed components of this framework is illustrated in Figure~\ref{fig:encoder-decoder}, where the ``Intermediate Bridge'' is responsible for implementing feature-related operations.%

As illustrated in Figure~\ref{fig:encoder-decoder}, the encoder extracts visual features $\mathbf{X}^{l}$ from the layer $l$ of the designated network and passes these features forward to the Intermediate Bridge, the output of which is then fed to the decoder (i.e., the description generator) to predict the subsequent word. The operations carried out inside the Intermediate Bridge, i.e., $\mathcal{B}_{\phi}\left(\mathbf{X}^{l}, \mathbf{W}\right)$, address the semantic gap between low-level encoder features and high-level linguistic context. The sequence of operations in Figure~\ref{fig:encoder-decoder} can be formalized as:%
\begin{equation}
    \mathbf{X}^{l} = \text{Encoder}\left(\mathbf{I}\right) \Rightarrow \mathbf{D} = \mathcal{B}_{\phi}\left(\mathbf{X}^{l}, \mathbf{W}\right) = \text{Concat}\left(\tilde{\mathbf{X}}, \tilde{\mathbf{W}}\right) \Rightarrow y_{s} = \text{Decoder}\left(\mathbf{D}\right)\text{.}%
\end{equation}%

\subsubsection{Feature Preservation}%
Since the introduction of Vision Transformer (ViT) \cite{2021_Dosovitskiy_ViT}, there has been a growing trend toward hybrid CNN-ViT architectures that combine both approaches for several compelling reasons, especially in AD, where many of these architectures apply CNNs as backbones for ViTs \cite{2024_Lai-Dang_Survey}. Pure ViT processes images by dividing them into patches and treating these patches as sequences, without employing any convolutional operations. However, hybrid architectures that incorporate CNN components have emerged as a popular approach because they address several limitations of pure ViT, such as data efficiency, training stability, complementary strengths, multi-scale feature processing, and computational efficiency \cite{2024_Yunusa_Survey}.%

In light of this, the present study proposes that feature hierarchies in hybrid CNN-ViT architectures affect the quality of description generation differently compared with classification and detection tasks. This hypothesis pertains to the encoder side of the encoder-decoder architecture, which is responsible for extracting features from a given input image. Essentially, this hypothesis posits that since earlier layers in the encoder contain more spatially relevant information than specialized later layers (i.e., the spatial information is concerned with local structure and position), spatial features can have a more positive impact on description generation; in other words, spatial feature preservation is more critical than semantic abstraction, which is concerned with object identity and global context, for accurate generation of scene descriptions. The effectiveness of features for this task can be formalized through the proposed utility function as:%
\begin{equation}
    \text{Utility}_{\text{description}}\left(\mathbf{X}^{l}\right) = \alpha_{l} \cdot I_{\text{spatial}}\left(\mathbf{X}^{l}\right) + \beta_{l} \cdot I_{\text{semantic}}\left(\mathbf{X}^{l}\right)\text{,}%
\end{equation}%
where $\text{Utility}_{\text{description}}\left(\mathbf{X}^{l}\right)$ stands for the utility or effectiveness of the extracted features $\mathbf{X}^{l}$ at layer $l$ for generating contextually accurate description of a gives scene, while $I_{\text{spatial}}\left(\mathbf{X}^{l}\right)$ and $I_{\text{semantic}}\left(\mathbf{X}^{l}\right)$ stand for the spatial and semantic information content of the extracted features; $\alpha_{l}$ and $\beta_{l}$ are importance weights at each layer.%

It is well-established that $\alpha_{l} > \beta_{l}$ in earlier layers and $\beta_{l} > \alpha_{l}$ in later layers. In other words, as the features are processed through different layers, they lose fine-grained spatial details ($\partial I_{\text{spatial}} / \partial l < 0$) due to various pooling and convolution operations, while gaining more high-level abstractions ($\partial I_{\text{semantic}}/ \partial l > 0$) in later layers. Therefore, this study proposes that when $\alpha_{l} > \beta_{l}$ for extracted features, the generated descriptions are more accurate; this notion can also be expressed as:%
\begin{equation}
    \frac{\partial \text{Utility}_{\text{description}}\left(\mathbf{X}^{l}\right)}{\partial I_{\text{spatial}}\left(\mathbf{X}^{l}\right)} > \frac{\partial \text{Utility}_{\text{description}}\left(\mathbf{X}^{l}\right)}{\partial I_{\text{semantic}}\left(\mathbf{X}^{l}\right)}\text{.}%
\end{equation}%

To evaluate this hypothesis, five encoder architectures are considered: two CNN-based models, two hybrid CNN-ViT-based models, and one pure ViT-based model. The CNN-based models are VGG-16 \cite{2014_Simonyan_VGG} and ResNet-50 \cite{2016_He_ResNet}. The hybrid models are the DEtection TRansformer (DETR) \cite{2020_Carion_DETR} and Real-Time DETR (RT-DETR) \cite{2024_Zhao_RT-DETR}. The ViT-based model is the Multiscale Vision Transformer (MViTv2) \cite{2022_Li_MViTv2}.%

DETR represents one of the first applications of Transformers in object detection, eliminating traditional components such as non-maximum suppression (NMS) and anchor generation. It leverages ResNet \cite{2016_He_ResNet} as its backbone for feature extraction. However, DETR faces challenges in detecting small objects due to its reliance on global features, which can limit its ability to capture fine-grained local details. Similarly, RT-DETR leverages ResNet \cite{2016_He_ResNet} as its backbone for feature extraction and continues to face challenges in detecting small objects and handling scenarios with dense occlusions. RT-DETR is the first real-time end-to-end object detector that aims to address limitations of YOLO models (i.e., dependency on NMS and computational overhead) and DETR models (i.e., high computational cost and inefficiency in real-time object detection). On the other hand, MViTv2 is an enhanced version of the MViT \cite{2021_Fan_MViT}, designed as a unified architecture for image and video classification, as well as object detection. It incorporates two key improvements: (i) decomposed relative positional embeddings to enhance shift invariance and (ii) residual pooling connections to mitigate the effects of pooling strides in attention computation. One of the main features of MViTv2 is the integration of pooling attention mechanisms to optimize accuracy and computational efficiency, outperforming local window attention methods, such as Swin Transformers \cite{2021_Liu_Swin_Transformer}, in various benchmarks.%

\subsubsection{Memory Enhancement}%
Numerous prior studies in the context of image captioning exploited LSTM \cite{1997_Hochreiter_LSTM} as the decoder to generate textual descriptions. While still popular and effective in processing sequential data and capturing long-term dependencies, LSTM exhibits limitations in capturing complex dependencies and generating long descriptions. On the other hand, xLSTM \cite{2024_Beck_xLSTM} is designed to scale LSTM and improve language modeling by mitigating the shortcomings of LSTM. This is accomplished by developing two new LSTM variants, scalar LSTM (sLSTM) and matrix LSTM (mLSTM), residually stacked within the xLSTM block. Since generating complex, detailed textual descriptions is crucial in traffic scene understanding, this study proposes exploiting xLSTM as the decoder for description generation. This is supported by the fact that the stabilized exponential gating technique%
\begin{equation}
    \mathbf{i}_{\text{f}}^{\prime} = \exp\left(\log\left(\mathbf{i}_{t}\right) - \max\left(\log\left(\mathbf{f}_{t}\right) + \mathbf{m}_{t - 1}, \log\left(\mathbf{i}_{t}\right)\right)\right)\text{,}%
\end{equation}%
as well as the new memory mixing techniques in sLSTM, enable the revision of the storage decision. The mSLTM, on the other hand, leverages a matrix-based memory structure with a covariance update rule%
\begin{equation}
    \mathbf{C}_{t} = \mathbf{f}_{t} \mathbf{C}_{t - 1} + \mathbf{i}_{t} \mathbf{v}_{t} \mathbf{k}_{t}^{\text{T}}\text{,}%
\end{equation}%
which stores key-value pairs, thus enabling more copious storage and fully parallelizable operations. Empirical evaluations demonstrate that xLSTM outperforms Transformers and State-Space models in terms of performance and scalability \cite{2024_Beck_xLSTM}.%

\subsubsection{Hybrid Attention Fusion}%
The attention layer takes as input three matrices of queries ($\mathbf{Q}$), keys ($\mathbf{K}$), and values ($\mathbf{V}$), and performs the following operation:%
\begin{equation}
    \text{Attention}\left(\mathbf{Q}, \mathbf{K}, \mathbf{V}\right) = \left[\text{Softmax}\left(\frac{\left(\mathbf{Q} \mathbf{W}^{\text{Q}}\right) \left(\mathbf{K} \mathbf{W}^{\text{K}}\right)^{\text{T}}}{\sqrt{d_{k}}}\right)\right] \left(\mathbf{V} \mathbf{W}^{\text{V}}\right)\text{,}%
\end{equation}%
where $\mathbf{W}^{\text{Q}}$, $\mathbf{W}^{\text{K}}$, and $\mathbf{W}^{\text{V}}$ are called projection matrices. In the following, three types of attention integration strategies are proposed to improve the quality of the generated description through feature enrichment.%

\textbf{Intra-Encoder Self-Attention Fusion.} To enrich the extracted features $\mathbf{X}^{l}$ at layer $l$ of the encoder, this study proposes the incorporation of a self-attention mechanism as the last layer of the encoder. This is based on the fact that self-attention enables each position in the feature map to attend to all other positions, thereby creating a more context-aware representation that enhances contextual relationships with single-modality features. The self-attention operation is as:%
\begin{equation}\label{SA}
    \mathbf{X}_{\text{SA}} = \text{SA}\left(\mathbf{X}^{l}, \mathbf{X}^{l} , \mathbf{X}^{l}\right) = \left[\text{Softmax}\left(\frac{\left(\mathbf{X}^{l} \mathbf{W}^{\text{Q}}\right) \left(\mathbf{X}^{l} \mathbf{W}^{\text{K}}\right)^{\text{T}}}{\sqrt{d_{k}}}\right)\right] \left(\mathbf{X}^{l} \mathbf{W}^{\text{V}}\right)\text{,}%
\end{equation}%
which is basically a projection of the query $\mathbf{X}^{l}$ onto the key $\mathbf{X}^{l}$ in a high-dimensional feature space, thus creating a non-local operation that outputs each position as a weighted sum of features at all positions. This ensures that the model optimally attends across all positions in the feature space, establishing long-term dependencies.%

\textbf{Inter-Encoder Cross-Attention Fusion.} For the same reason of feature enrichment, this study proposes two variations of cross-attention: (i) unidirectional cross-attention fusion and (ii) bidirectional cross-attention fusion. Here, the cross-attention mechanism is performed between two encoders; the rationale behind adding another encoder is to enable a different feature extraction stream that captures different aspects of the same input data. This dual-encoder approach enables each encoder to develop expertise in distinct aspects of the spatial feature space, and the cross-attention mechanism facilitates the exchange of information between these complementary representations. The cross-attention operation is as:%
\begin{equation}
    \mathbf{X}_{\text{UCAF}} = \text{CA}\left(\mathbf{X}_{\text{A}}^{l_{\text{A}}}, \mathbf{X}_{\text{B}}^{l_{\text{B}}}, \mathbf{X}_{\text{B}}^{l_{\text{B}}}\right) = \left[\text{Softmax}\left(\frac{\left(\mathbf{X}_{\text{A}}^{l_{\text{A}}} \mathbf{W}^{\text{Q}}\right) \left(\mathbf{X}_{\text{B}}^{l_{\text{B}}} \mathbf{W}^{\text{K}}\right)^{\text{T}}}{\sqrt{d_{k}}}\right)\right] \left(\mathbf{X}_{\text{B}}^{l_{\text{B}}} \mathbf{W}^{\text{V}}\right)\text{,}%
\end{equation}%
where $\mathbf{X}_{\text{A}}^{l_{\text{A}}}$ represents the extracted features at layer $l_{\text{A}}$ of the encoder A and $\mathbf{X}_{\text{B}}^{l_{\text{B}}}$ represents the extracted features at layer $l_{\text{B}}$ of the encoder B. This operation is also called the unidirectional cross-attention fusion, hence $\mathbf{X}_{\text{UCAF}}$. On the other hand, bidirectional cross-attention fusion comprises two unidirectional cross-attention fusion operations; the results of both operations are then passed through a feedforward neural network (FFNN) layer to transform them so that their dimensions are the same. Finally, a weighted averaging operator is leveraged to fuse the features as:%
\begin{equation}
    \mathbf{X}_{\text{BCAF}} = \lambda \cdot \text{FFNN}\left(\text{CA}\left(\mathbf{X}_{\text{A}}^{l_{\text{A}}}, \mathbf{X}_{\text{B}}^{l_{\text{B}}}, \mathbf{X}_{\text{B}}^{l_{\text{B}}}\right)\right) + \left(1 - \lambda\right) \cdot \text{FFNN}\left(\text{CA}\left(\mathbf{X}_{\text{B}}^{l_{\text{B}}}, \mathbf{X}_{\text{A}}^{l_{\text{A}}}, \mathbf{X}_{\text{A}}^{l_{\text{A}}}\right)\right)\text{,}%
\end{equation}%
where $\lambda$ is a learnable parameter. This operation is hypothesized to enable bidirectional information flow between complementary encoders, thereby enriching the extracted features.%

\textbf{Multimodal Cross-Attention Fusion.} Another usage of cross-attention is to bridge the semantic gap between visual modality and textual modality by aligning the textual description generation with visual features via dynamic feature space projection, as well as context-aware attention weighting as:%
\begin{equation}\label{mm}
    \tilde{\mathbf{W}} = \text{CA}\left(\mathbf{W}, \tilde{\mathbf{X}}, \tilde{\mathbf{X}}\right) = \left[\text{Softmax}\left(\frac{\left(\mathbf{W} \mathbf{W}^{\text{Q}}\right) \left(\tilde{\mathbf{X}} \mathbf{W}^{\text{K}}\right)^{\text{T}}}{\sqrt{d_{k}}}\right)\right] \left(\tilde{\mathbf{X}} \mathbf{W}^{\text{V}}\right)\text{,}%
\end{equation}%
where $\mathbf{W}$ stands for the word embeddings of previously generated words $\mathbf{y}_{< s}$ during inference or the word embeddings of reference descriptions during training. Furthermore, $\tilde{\mathbf{X}}$ in \eqref{mm} can be $\mathbf{X}^{l}$, $\mathbf{X}_{\text{SA}}$, $\mathbf{X}_{\text{UCAF}}$, or $\mathbf{X}_{\text{BCAF}}$. This operation can be interpreted as attention of textual queries to visual keys and values, thereby refining the textual information through cross-modal alignment. The multimodal cross-attention fusion is then followed by the decoding step (assuming $\mathbf{W}$ is the word embeddings of previously generated words $\mathbf{y}_{< s}$) as:%
\begin{equation}
    y_{s} = \text{Decoder}\left(\text{Concat}\left(\tilde{\mathbf{X}}, \tilde{\mathbf{W}}\right)\right)\text{.}%
\end{equation}%

\textbf{Static Image Temporalization.} To investigate the effect of features extracted by a video encoder from input images, this study proposes the transformation of static images into temporal sequences as:%
\begin{equation}
    \mathbf{I}_{\text{temporal}} = \text{Repeat}\left(\mathbf{I}, T\right)\text{,}%
\end{equation}%
where $T$ is the number of frames, $\mathbf{I}$ is the input image, and function $\text{Repeat}\left(\cdot\right): \mathbb{R}^{C \times H \times W} \rightarrow \mathbb{R}^{C \times T \times H \times W}$ is a pseudo-temporal sequence constructor, stacking $T$ copies of image $\mathbf{I}$ to create a pseudo-temporal representation of a single input image for the video encoder. The underlying reason for adopting this approach is that the temporal modeling capabilities of video encoders can identify features that remain invariant across the temporal dimension and become spatial feature enhancers, thereby enriching the spatial features.%

\subsection{Traffic Scene Dataset}%
As previously mentioned, dataset inaccessibility and detail insufficiency are among the main issues in traffic scene understanding through natural language text generation within the context of AD. Therefore, the present study proposes creating an appropriate dataset for this purpose. This is achieved by selecting a suitable camera-based autonomous driving dataset and writing descriptive sentences for the images. To create an appropriate dataset for scene understanding, several key factors must be considered \cite{2023_Aldoski,2023_Zhang_Survey,2024_Zhang_ESAVAWDLBOD}, as outlined below:%
\begin{itemize}
    \item \textbf{Regulatory signs}---including speed limits, stop signs, yield signs, and no-entry signs.%
    \item \textbf{Warning signs}---including sharp turns, pedestrian crossings, school zones, and slippery roads.%
    \item \textbf{Information signs}---including directions, distances to destinations, and locations of facilities such as hospitals and gas stations.%
    \item \textbf{Lane markings}---including number of lanes and lane usage (e.g., turn lanes).%
    \item \textbf{Road surface}---including any potholes, uneven surfaces, and construction zones.%
    \item \textbf{Intersections}---including traffic lights, stop signs, and type of intersection (e.g., roundabouts, T-junctions).%
    \item \textbf{Pedestrians}---including presence of pedestrians, crosswalks, pedestrian signals, school zones, and shopping areas.%
    \item \textbf{Cyclists}---including presence of cyclists, bike lanes, and shared road signs.%
    \item \textbf{Vehicles}---including presence of other vehicles and their types (e.g., cars, buses, or trucks).%
    \item \textbf{Depth and distance estimation}---including estimated distance to various objects and obstacles.%
    \item \textbf{Weather}---including weather conditions such as rain, fog, or snow.%
    \item \textbf{Lighting}---including time of day and lighting conditions, such as day or night.%
\end{itemize}%

Following the above factors and incorporating them in the human-generated reference scene descriptions leads to a safe, efficient, and reliable driving experience; allows the ego vehicle to understand the environment; and helps AVs (and drivers, in some cases) make informed decisions \cite{2023_Aldoski,2023_Zhang_Survey,2024_Zhang_ESAVAWDLBOD}. The information can also be shared with other vehicles to elevate the safety levels of the driving experience in car platoons.%

In addition to the factors mentioned above related to the safe driving aspect of the scene descriptions, there are other guidelines to follow when writing human-generated reference scene descriptions. For example, in \cite{2015_Chen_MS_COCO_Captions}, the authors created captions/descriptions for the {MS~COCO} dataset \cite{2014_Lin_MS_COCO} through the Amazon Mechanical Turk platform by crowdsourcing the creation of reference captions. Through a user interface, human contributors (or data annotators) were presented with an image, for which they were asked to write a caption, along with a set of rules to follow when writing captions. This set of rules is not presented here for the sake of brevity, and the interested reader is encouraged to refer to \cite{2015_Chen_MS_COCO_Captions} for more information. Furthermore, according to \cite{2022_Kasai,2023_Leotta}, other factors must be considered when creating human-generated reference captions. These factors, adapted to AD for traffic scene understanding, are outlined below:%
\begin{enumerate}
    \item \textbf{Semantic Accuracy:} Descriptions are required to be semantically accurate when describing various aspects of a scene, such as various entities or objects, their positions with respect to the ego vehicle, and maybe also their actions.%
    \item \textbf{Conciseness and Precision:} Descriptions are required to deliver important information about the scene while avoiding verbosity or odd phrases.%
    \item \textbf{Saliency Coverage:} Descriptions are required to cover the most salient features and most critical details of the scene.%
    \item \textbf{Fluency and Readability:} Descriptions, regardless of the content of the scene they are describing, are required to be clear, understandable, and grammatically correct.%
    \item \textbf{Inclusivity and Sensitivity:} Descriptions are required to represent and respect all groups of people, avoid stereotypes and biases, and steer clear of severe or subjective comments.%
    \item \textbf{Alignment with Human Driver's Perception:} Descriptions are required to mirror the perception of a human driver, reflecting the way human drivers see and interpret the driving environment.%
\end{enumerate}%

Considering all factors and rules mentioned above, the present study proposes a novel set of guidelines, presented in~\ref{app-GDA}, for dataset annotation. Unlike previous studies, which provided one or five sentences for each image in their dataset, the current research associates ten sentences to each image in the selected dataset, as it has been observed that the correlation between evaluation metrics and human judgment is increased when more reference sentences are provided \cite{2015_Vedantam_CIDEr,2015_Chen_MS_COCO_Captions}.%

\subsubsection{Selected Image Dataset}%
The dataset selected for this study is BDD100K \cite{2020_Yu_BDD100K}, primarily a driving video dataset comprising 100,000 videos and 10 tasks. This dataset was collected to assess the advancement of image recognition algorithms in the field of autonomous driving. The dataset encompasses a wide range of geographical, ecological, and climatic attributes, exhibiting significant diversity in these domains. In addition to the videos, the dataset includes two subsets: (i) 100K Images and (ii) 10K Images. The present study selects the 10K Images subset to investigate the proposed framework and develop an image-text dataset for traffic scene understanding. Basically, the 10K Images subset is used for semantic, instance, and panoptic segmentation, and it significantly overlaps with the 100K Images subset.%

\subsubsection{Data Annotation}%
A subset of 600 images was selected from the 10K Images dataset, and the selected images were then manually annotated with appropriate traffic scene descriptions based on the proposed annotation guidelines in~\ref{app-GDA}. The annotation task was outsourced to three individuals with driving experience, who were instructed to write ten descriptive sentences for each image while adhering to guidelines established by the researcher. The researcher then thoroughly checked and reviewed the human-generated descriptions to ensure they adhered to the established guidelines. Table~\ref{tab:developed-dataset-p01} and Table~\ref{tab:developed-dataset-p02} depict some sample images from the 10K Images dataset and their corresponding human-generated reference descriptions. As presented in these tables, at least one sentence explicitly states the weather conditions (clear, rainy, snowy, foggy) and lighting conditions (daytime or nighttime). For example, ``\texttt{It is clear daytime.}'' or ``\texttt{It is clear nighttime.}'' or ``\texttt{It is daytime/nighttime, and it is snowing/raining.}'' The descriptive sentences in the developed dataset vary in length; some are short to allow models to concentrate on one or two critical aspects at a time, while others are long, enabling the model to generate complex, lengthy descriptions.%

\clearpage%
\newpage%
\newgeometry{left=0.2in, right=0.2in, top=0.2in, bottom=0.2in}%
\begin{table}[!t]
    \centering%
    \begin{tabularx}{\textwidth}{lX}
        \toprule%
        \makecell[c]{\textit{seen\_bdd\_001}\\{\includegraphics[height=1.00in]{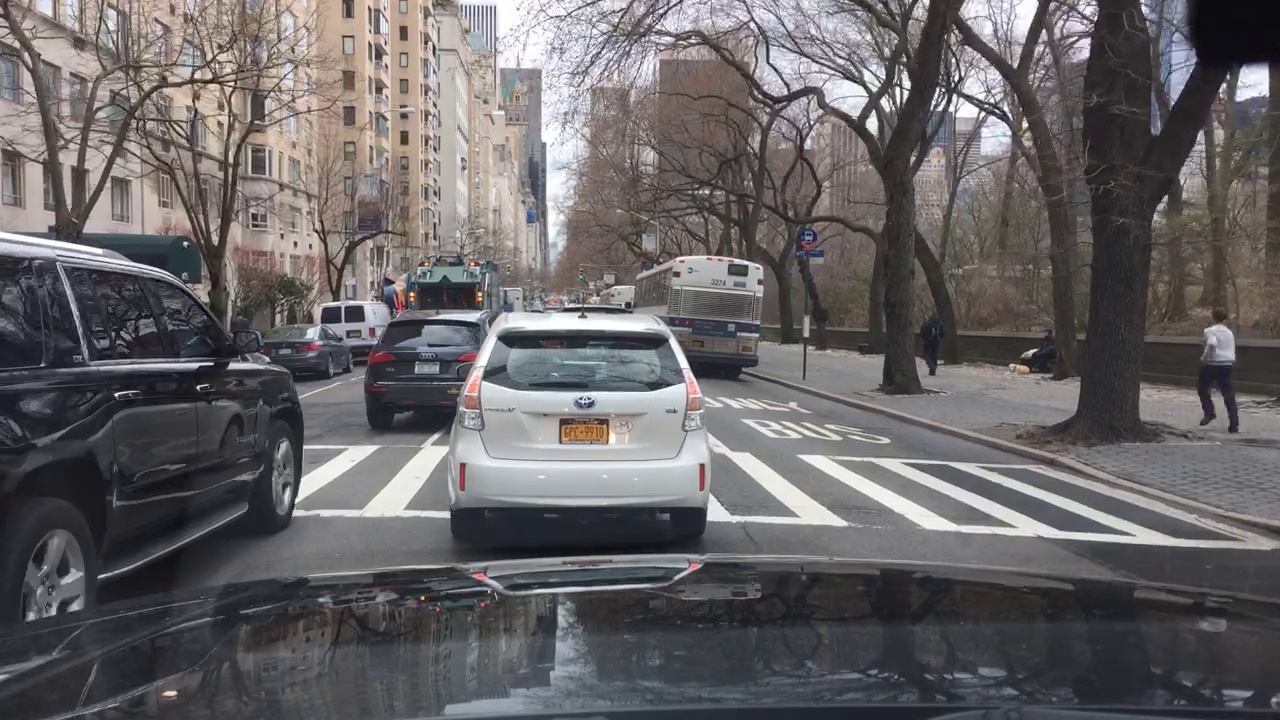}}} & \makecell[X]{\scriptsize{\texttt{\fbox{C01:} It is clear daytime.}}\\%
        \scriptsize{\texttt{\fbox{C02:} It is a multi-lane street.}}\\%
        \scriptsize{\texttt{\fbox{C03:} A white car is driving in the ego lane nearby.}}\\%
        \scriptsize{\texttt{\fbox{C04:} It is a residential area.}}\\%
        \scriptsize{\texttt{\fbox{C05:} A crosswalk is ahead, and 1 white car is driving in the ego lane nearby.}}\\%
        \scriptsize{\texttt{\fbox{C06:} No pedestrians are on the crosswalk.}}\\%
        \scriptsize{\texttt{\fbox{C07:} 3 people are on the right sidewalk, and the right lane is a bus lane.}}\\%
        \scriptsize{\texttt{\fbox{C08:} The right lane is a bus lane, and there is a bus in the right lane.}}}\\%
        \multicolumn{2}{p{\textwidth}}{\scriptsize{\texttt{\fbox{C09:} Many cars are on the street, and the right lane is a bus lane.}}}\\%
        \multicolumn{2}{p{\textwidth}}{\scriptsize{\texttt{\fbox{C10:} It is clear daytime, it is a multi-lane street, a crosswalk is ahead, one white car is driving in the ego lane nearby, no pedestrians are on the crosswalk, many cars are on the street, and it is a residential area.}}}\\%
        \midrule%
        \makecell[c]{\textit{seen\_bdd\_002}\\{\includegraphics[height=1.00in]{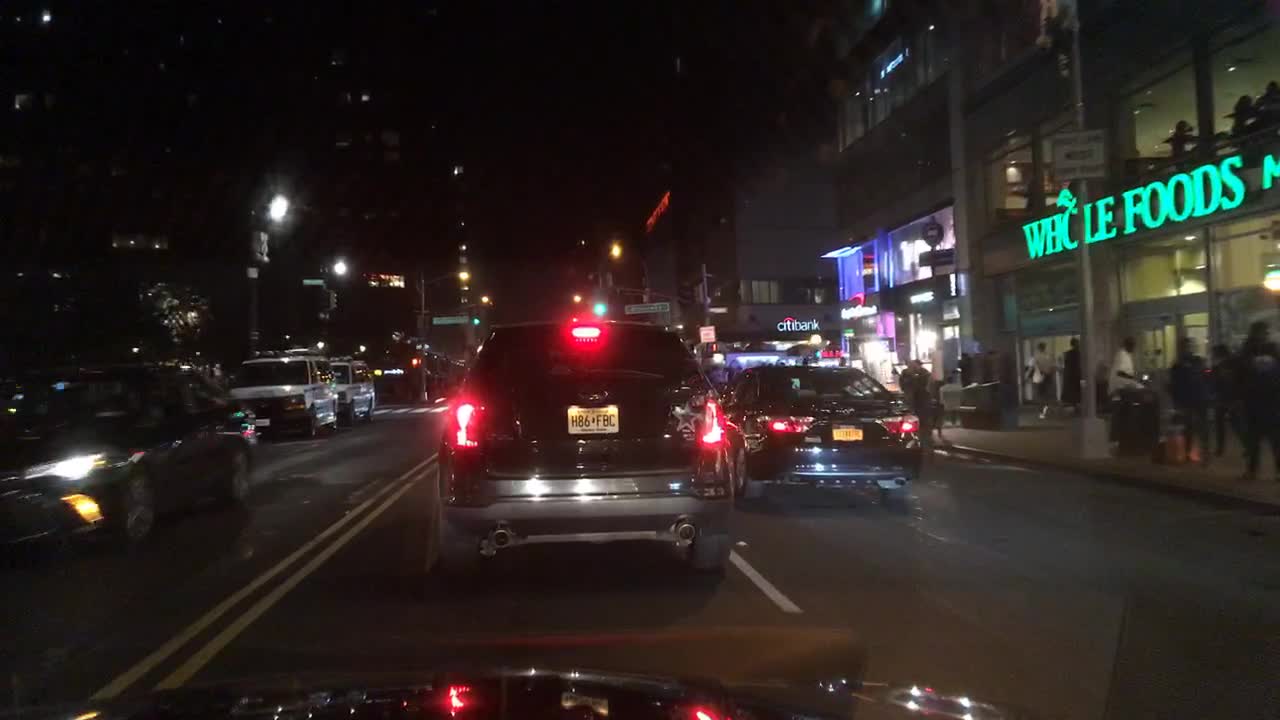}}} & \makecell[X]{\scriptsize{\texttt{\fbox{C01:} It is clear nighttime.}}\\%
        \scriptsize{\texttt{\fbox{C02:} It is a two-way street.}}\\%
        \scriptsize{\texttt{\fbox{C03:} A black SUV is braking in front with its brake lights on.}}\\%
        \scriptsize{\texttt{\fbox{C04:} 1 black SUV is braking in front with its brake lights on.}}\\%
        \scriptsize{\texttt{\fbox{C05:} Many pedestrians are walking on the right sidewalk.}}\\%
        \scriptsize{\texttt{\fbox{C06:} An intersection is ahead.}}\\%
        \scriptsize{\texttt{\fbox{C07:} The traffic light is green at the intersection.}}\\%
        \scriptsize{\texttt{\fbox{C08:} The ego lane is the leftmost lane, and 1 black SUV is braking in front with its brake lights on.}}}\\%
        \multicolumn{2}{p{\textwidth}}{\scriptsize{\texttt{\fbox{C09:} It is clear nighttime, it is a two-way street, one black SUV is braking in front with its brake lights on, the ego lane is the leftmost lane, an intersection is ahead, and the traffic light is green at the intersection.}}}\\%
        \multicolumn{2}{p{\textwidth}}{\scriptsize{\texttt{\fbox{C10:} It is clear nighttime, it is a two-way street, and 1 black SUV is braking in front with its brake lights on.}}}\\%
        \midrule%
        \makecell[c]{\textit{seen\_bdd\_003}\\{\includegraphics[height=1.00in]{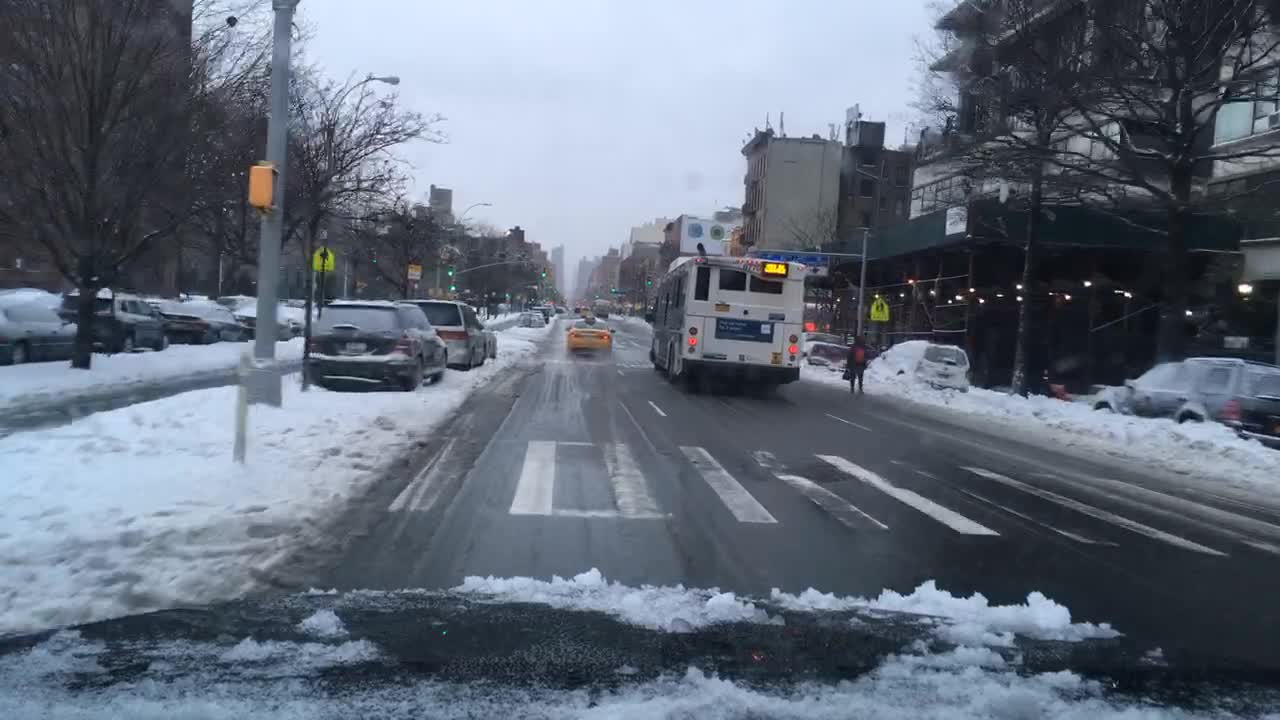}}} & \makecell[X]{\scriptsize{\texttt{\fbox{C01:} It is daytime, and the street is wet and slippery with snow.}}\\%
        \scriptsize{\texttt{\fbox{C02:} It is a two-lane street, and a crosswalk is ahead in front.}}\\%
        \scriptsize{\texttt{\fbox{C03:} There is snow on the street, and the street is wet and slippery with snow.}}\\%
        \scriptsize{\texttt{\fbox{C04:} A yellow taxi is driving in the ego lane some distance ahead.}}\\%
        \scriptsize{\texttt{\fbox{C05:} 2 cars are parked on the left side of the road.}}\\%
        \scriptsize{\texttt{\fbox{C06:} A pedestrian is on the right side of the road.}}\\%
        \scriptsize{\texttt{\fbox{C07:} A bus is braking in the right lane with its brake lights on.}}\\%
        \scriptsize{\texttt{\fbox{C08:} The ego lane is the leftmost lane, and a crosswalk is ahead in front.}}}\\%
        \multicolumn{2}{p{\textwidth}}{\scriptsize{\texttt{\fbox{C09:} A [SCHOOL ZONE] sign is on the right side of the street ahead.}}}\\%
        \multicolumn{2}{p{\textwidth}}{\scriptsize{\texttt{\fbox{C10:} It is clear daytime, the street is wet and slippery with snow, it is a two-lane street, a taxi is driving in the ego lane ahead, 2 cars are parked on the left side of the road, a pedestrian is on the right side of the road, 1 bus is braking in the right lane with its brake lights on, and there is a [SCHOOL ZONE] sign on the right side of the street ahead.}}}\\%
        \bottomrule%
    \end{tabularx}%
    \caption{Sample Images from 10K Images Dataset \cite{2020_Yu_BDD100K} and Their Corresponding Human-Generated Reference Descriptions (Developed in the Present Study) - Part 1}\label{tab:developed-dataset-p01}%
\end{table}%
\thispagestyle{empty}%
\restoregeometry
\newpage%
\clearpage
\newpage%

\clearpage%
\newpage%
\newgeometry{left=0.2in, right=0.2in, top=0.2in, bottom=0.2in}%
\begin{table}[!t]
    \centering%
    \begin{tabularx}{\textwidth}{lX}
        \toprule%
        \makecell[c]{\textit{seen\_bdd\_004}\\{\includegraphics[height=1.00in]{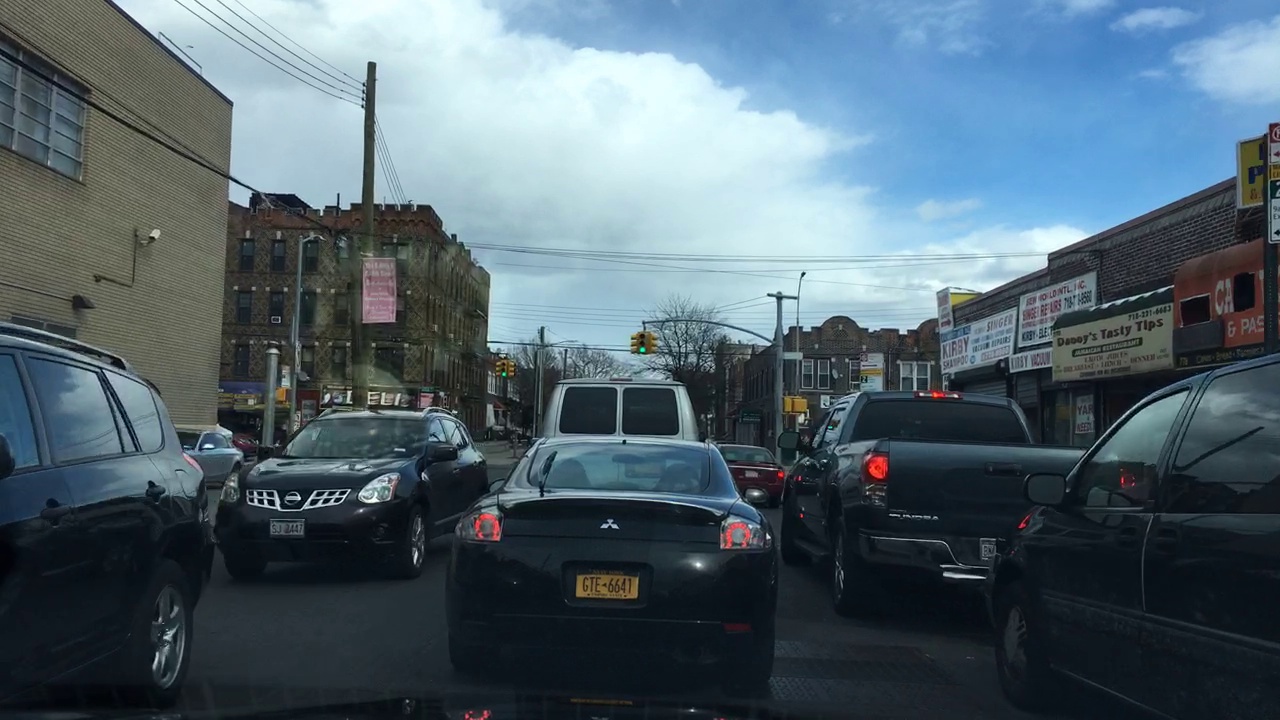}}} & \makecell[X]{\scriptsize{\texttt{\fbox{C01:} It is clear daytime.}}\\%
        \scriptsize{\texttt{\fbox{C02:} It is a two-way street.}}\\%
        \scriptsize{\texttt{\fbox{C03:} There is traffic congestion ahead.}}\\%
        \scriptsize{\texttt{\fbox{C04:} 1 black car with its brake lights on is braking nearby in front.}}\\%
        \scriptsize{\texttt{\fbox{C05:} One black truck with its brake lights on is braking in the right lane.}}\\%
        \scriptsize{\texttt{\fbox{C06:} Many vehicles with their brake lights on are on the road.}}\\%
        \scriptsize{\texttt{\fbox{C07:} A traffic light is at intersection in front, and the traffic light is green in front.}}\\%
        \scriptsize{\texttt{\fbox{C08:} The traffic light is green in front.}}}\\%
        \multicolumn{2}{p{\textwidth}}{\scriptsize{\texttt{\fbox{C09:} 2 black cars are in the left lane in the opposite direction.}}}\\%
        \multicolumn{2}{p{\textwidth}}{\scriptsize{\texttt{\fbox{C10:} It is clear daytime, there is traffic congestion ahead, many vehicles are braking with their brake lights on, the traffic light is green at intersection in front, 1 black car with its brake lights on is braking nearby in front in the ego lane, One black truck with its brake lights on is braking in the right lane, and 2 black cars are in the left lane in the opposite direction.}}}\\%
        \midrule%
        \makecell[c]{\textit{seen\_bdd\_005}\\{\includegraphics[height=1.00in]{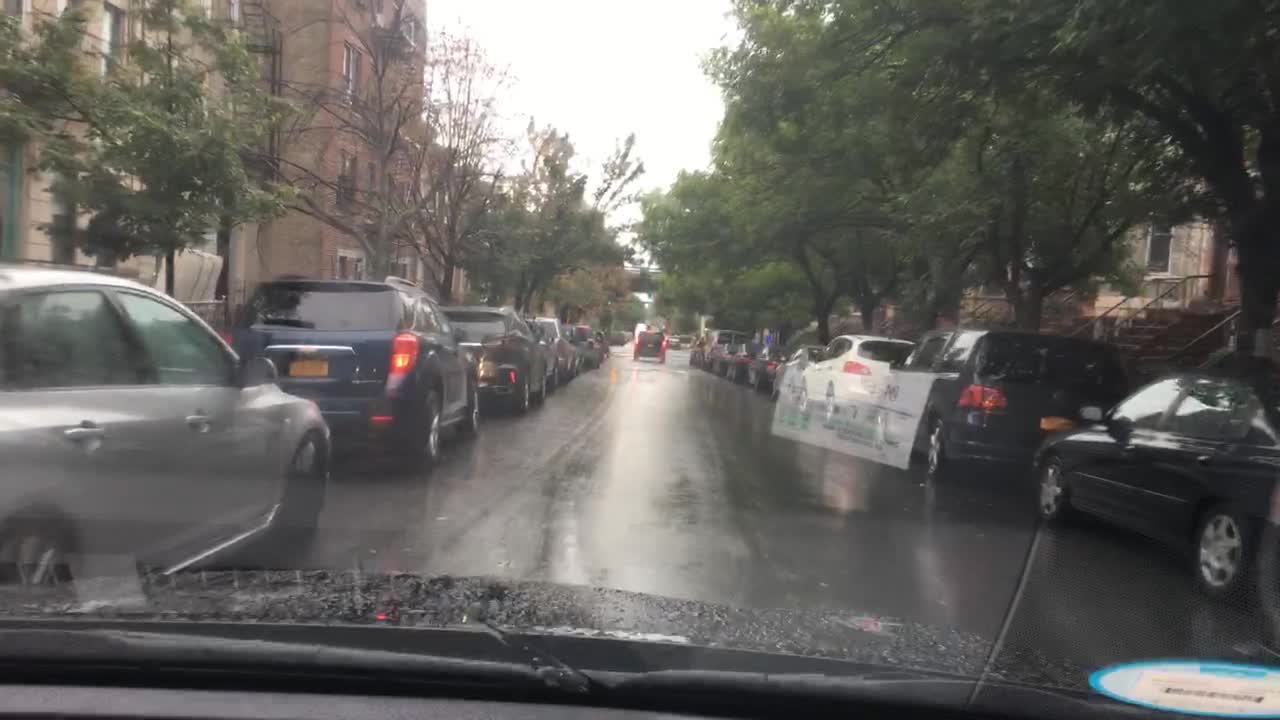}}} & \makecell[X]{\scriptsize{\texttt{\fbox{C01:} It is daytime, and it is raining.}}\\%
        \scriptsize{\texttt{\fbox{C02:} The road is wet and slippery because of the rain.}}\\%
        \scriptsize{\texttt{\fbox{C03:} It is a one-way, narrow street.}}\\%
        \scriptsize{\texttt{\fbox{C04:} It is a one-way, narrow street, and the street is wet and slippery because of the rain.}}\\%
        \scriptsize{\texttt{\fbox{C05:} 1 vehicle is braking in the ego lane some distance ahead with its taillights and brake lights on.}}\\%
        \scriptsize{\texttt{\fbox{C06:} Many vehicles are parked on both sides of the street.}}\\%
        \scriptsize{\texttt{\fbox{C07:} It is raining, and the road is wet and slippery because of the rain.}}}\\%
        \multicolumn{2}{p{\textwidth}}{\scriptsize{\texttt{\fbox{C08:} The road is wet and slippery because of the rain, and a vehicle is braking in the ego lane some distance ahead with its taillights and brake lights on.}}}\\%
        \multicolumn{2}{p{\textwidth}}{\scriptsize{\texttt{\fbox{C09:} It is a one-way, narrow street, the street is wet and slippery because of the rain, and Many vehicles are parked on both sides of the street.}}}\\%
        \multicolumn{2}{p{\textwidth}}{\scriptsize{\texttt{\fbox{C10:} It is daytime, it is raining, It is a one-way, narrow street, the street is wet and slippery because of the rain, Many vehicles are parked on both sides of the street, and a vehicle is braking in the ego lane some distance ahead with its taillights and brake lights on.}}}\\%
        \bottomrule%
    \end{tabularx}%
    \caption{Sample Images from 10K Images Dataset \cite{2020_Yu_BDD100K} and Their Corresponding Human-Generated Reference Descriptions (Developed in the Present Study) - Part 2}\label{tab:developed-dataset-p02}%
\end{table}%
\thispagestyle{empty}%
\restoregeometry
\newpage%
\clearpage
\newpage%

\subsection{Evaluation Metrics}%
This section provides a concise overview of the evaluation metrics commonly employed in image captioning research. While it is typically unnecessary to describe these metrics in detail, a brief introduction is warranted here, given the extensive discussion of metric-related findings presented in this study.%

\subsubsection{BLEU (BiLingual Evaluation Understudy)}%
Papineni et al. \cite{2002_Papineni_BLEU} proposed BLEU, an automated understudy to human judges, for quick and language-independent evaluation of machine translation. BLEU mainly compares the $n$-grams of the candidate sentence with the $n$-grams of the reference sentence and then counts the number of position-independent matches. To calculate the BLEU score, the first step is to calculate the precision for each $n$-gram as:%
\begin{equation}
    p_{n} = \frac{\sum_{C \in \{\text{Candidates}\}} \sum_{\text{$n$-gram} \in C} \text{Count}_{\text{clip}}\left(\text{$n$-gram}\right)}{\sum_{C^{\prime} \in \{\text{Candidates}\}} \sum_{\text{$n$-gram}^{\prime} \in C^{\prime}} \text{Count}\left(\text{$n$-gram}^{\prime}\right)}\text{,}
\end{equation}%
where $\text{\text{Count}}\left(\cdot\right)$ counts the number of times an $n$-gram appears in the candidate sentence and%
\begin{equation}
    \text{Count}_{\text{clip}}\left(\text{$n$-gram}\right) = \min\left(\text{Count}\left(\text{$n$-gram}\right), \text{Maximum\_Reference\_Count}\right)\text{.}%
\end{equation}%
$\text{Count}_{\text{clip}}\left(\cdot\right)$ computes precision via clipping, and it calculates precision for any word with respect to the maximum number of times the word appears in all reference sentences. Next, the candidate sentences shorter than the reference sentences must be penalized through the brevity penalty (PB) as:%
\begin{equation}
    \text{PB} =
    \begin{cases}
        1 & \text{if $c > r$}\\%
        \exp\left(1 - \frac{r}{c}\right) & \text{if $c \leq r$}%
    \end{cases}\text{,}%
\end{equation}%
where $c$ is the length of the candidate sentence, and $r$ is the effective reference length, calculated by summing the best match length for each candidate sentence. Finally, the BLEU score is calculated as:%
\begin{equation}\label{BLEU}
    \text{BLEU} = \text{BP} \cdot \exp\left(\sum_{n=1}^{N} w_{n} \cdot \ln\left(p_{n}\right)\right)\text{,}%
\end{equation}%
where $N$ is the maximum length of $n$-grams (typically $N = 4$) and $w_{n}$ are positive weights for each $n$-gram (typically $w_{n} = 1/N$).%

\subsubsection{ROUGE-L (Recall-Oriented Understudy for Gisting Evaluation)}%
Lin \cite{2004_Lin_ROUGE} introduced ROUGE as an automatic evaluation package for measuring the quality of text summaries through a comparison with other human-generated summaries as references. ROUGE-L measures the longest common subsequence (LCS) between a candidate summary sentence ($Y$) and a reference summary sentence ($X$). The longer the LCS, the more similar the two summaries. Let $m$ be the length of $X$, as reference text, and $n$ the length of $Y$, as candidate text; the F-measure based on LCS to calculate the similarity between sentences is as:%
\begin{equation}
    R_{\text{lcs}} = \frac{\text{LCS}\left(X, Y\right)}{m}\text{,~} P_{\text{lcs}} = \frac{\text{LCS}\left(X, Y\right)}{n}\text{,~} F_{\textit{lcs}} = \frac{\left(1 + \beta^{2}\right) \cdot R_{\text{lcs}} \cdot P_{\text{lcs}}}{R_{\text{lcs}} + \beta^{2} \cdot P_{\text{lcs}}}\text{,}%
\end{equation}%
where $\beta$ is a parameter that adjusts the balance between precision and recall. $\beta = 1$ means that precision and recall have equal weight.%

\subsubsection{METEOR (Metric for Evaluation of Translation with Explicit word ORdering)}%
METEOR metric was first introduced in \cite{2004_Lavie_First_METEOR} as an automatic machine translation evaluation metric that calculates a score by comparing the candidate translation to a reference translation with explicit word-for-word matching \cite{2007_Lavie_Third_METEOR}. More details can be found in \cite{2004_Lavie_First_METEOR,2005_Banerjee_Second_METEOR,2007_Lavie_Third_METEOR}. The calculation of the METEOR score is as:%
\begin{equation}
    R = \frac{\text{number of mapped unigrams between two strings}}{\text{total number of unigrams in reference}}\text{,}%
\end{equation}%
\begin{equation}
    P = \frac{\text{number of mapped unigrams between two strings}}{\text{total number of unigrams in translation}}\text{,}%
\end{equation}%
and%
\begin{equation}
    F_{\text{mean}} = \frac{P \cdot R}{\alpha \cdot P + \left(1 - \alpha\right) \cdot R}\text{.}%
\end{equation}%
Single-word matches are the basis of $F_{\text{mean}}$, $P$, and $R$. A penalty, $p$, is defined to consider the degree to which the unigrams in the candidate and reference sentences match in the same order. Initially, the sequences of matched unigrams are separated into the minimal quantity of ``chunks'' as possible to ensure that the matched unigrams in each chunk are not only adjacent in both sentences but are also in the same word order. This penalty is calculated as:%
\begin{equation}
    p = \gamma \cdot \left(\frac{\text{number of chunks}}{\text{number of matches}}\right)^{\beta}\text{,}%
\end{equation}%
where $\gamma \in [0, 1]$ controls the maximum penalty and $\beta$ controls the fractional relationship in the penalty. The METEOR score for the alignment between the sentences is calculated as:%
\begin{equation}
    \text{METEOR} = \left(1 - p\right) \cdot F_{\text{mean}}\text{.}%
\end{equation}%

\subsubsection{CIDEr (Consensus-based Image Description Evaluation)}%
Vedantan et al. \cite{2015_Vedantam_CIDEr} introduced CIDEr as an evaluation protocol based on human consensus, measuring the resemblance of a candidate sentence to the consensus of human-generated sentences. CIDEr, through sentence resemblance, intrinsically captures the concepts of grammar, importance, accuracy, and salience. Let $c_{i}$ be a candidate sentence for image $I_{i}$ and $S_{i} = \{s_{i1}, \dots, s_{im}\}$ be a set of human-generated or reference descriptions for that image; CIDEr score is calculated as:%
\begin{equation}
    \text{CIDEr}\left(c_{i}, S_{i}\right) = \frac{1}{m} \sum_{n=1}^{N} \sum_{j=1}^{m} w_{n} \cdot \frac{\mathbf{g^{n}}\left(c_{i}\right) \mathbf{g^{n}}(s_{ij})}{\lVert \mathbf{g^{n}}\left(c_{i}\right) \rVert \lVert \mathbf{g^{n}}\left(s_{ij}\right) \rVert}\text{,}%
\end{equation}%
where $\mathbf{g^{n}}\left(s_{ij}\right)$ is a vector created by $g_{k}\left(s_{ij}\right)$ in accordance with all $n$-grams of length $n$ and $\lVert \mathbf{g^{n}}\left(s_{ij}\right) \rVert$ is the magnitude of $\mathbf{g^{n}}\left(s_{ij}\right)$. For the weights, the authors stated that uniform weights of $w_{n} = 1/N$, where $N = 4$, work best. Furthermore, $g_{k}\left(s_{ij}\right)$ is calculated as:%
\begin{equation}
    g_{k}\left(s_{ij}\right) = \frac{h_{k}\left(s_{ij}\right)}{\sum_{v_{l} \in \Omega} h_{l}\left(s_{ij}\right)} \log\left(\frac{\lvert I \rvert}{\sum_{I_{p} \in I} \min\left(1, \sum_{q} h_{k}\left(s_{pq}\right)\right)}\right)\text{,}%
\end{equation}%
where $v_{n}$ is an $n$-gram consisting of a set of one or more ordered words, $h_{k}\left(s_{ij}\right)$ is the number of times $v_{n}$ occurs in $s_{ij}$ for the reference sentence, and $\Omega$ is the vocabulary of all $n$-grams. The same procedure holds for $\mathbf{g^{n}}\left(c_{i}\right)$ and $g_{k}\left(c_{i}\right)$. $g_{k}(\cdot)$ is called the Term Frequency-Inverse Document Frequency (TF-IDF) weight for each $n$-gram \cite{2004_Robertson_TF-IDF}; the reason is to consider the relative value of various $n$-grams and to highlight those that are less common and more informative throughout the dataset.%

\subsubsection{SPICE (Semantic Propositional Image Caption Evaluation)}%
Anderson et al. \cite{2016_Anderson_SPICE} introduced SPICE to overcome the shortcomings of the previous metrics, as they are sensitive to $n$-gram overlap; this is because $n$-gram overlap does not necessarily mean that two sentences have the same meaning \cite{2007_Gimenez}. SPICE is proposed based on scene graphs, as the authors hypothesize that semantic propositional contents play a pivotal role in human judgment. The scene graph is used to represent semantic meaning with the purpose of encoding the semantic propositional content of the captions. Let $c$ be a candidate sentence for image $I$, $C$ be a set of object classes, $S = \{s_{1}, \dots, s_{m}\}$ be a set of human-generated or reference captions for image $I$, $R$ be a set of relation types, $A$ be a set of attribute types, and $G(c)$ and $G(S)$ be scene graphs of captions; the generation of scene graphs by parsing captions is as:%
\begin{equation}
    G(c) = \langle O(c), E(c), K(c) \rangle\text{,}%
\end{equation}%
where $O(c) \subseteq C$ represents the set of objects in $c$, $E(c) \subseteq O(c) \times R \times O(c)$ represents the set of hyper-edges showcasing relations between objects, and $K(c) \subseteq O(c) \times A$ represents the set of attributes corresponding to objects. The similarity between the candidate and reference scene graphs is evaluated by the F1-score as:%
\begin{equation}
    P\left(c, S\right) = \frac{\lvert T\left(G(c)\right) \otimes T\left(G(S)\right) \rvert}{\lvert T\left(G(c)\right) \rvert}\text{,~} R\left(c, S\right) = \frac{\lvert T\left(G(c)\right) \otimes T\left(G(S)\right) \rvert}{\lvert T\left(G(S)\right) \rvert}\text{,}%
\end{equation}%
\begin{equation}
    \text{SPICE}\left(c, S\right) = F1\left(c, S\right) = \frac{2 \cdot P\left(c, S\right) \cdot R\left(c, S\right)}{P\left(c, S\right) + R\left(c, S\right)}\text{,}%
\end{equation}%
where $T\left(G(c)\right) \triangleq O(c) \cup E(c) \cup K(c)$ is a function that outputs logical tuples from a scene graph, and $\otimes$ is a binary matching operator that outputs the matching tuples in candidate and reference scene graphs. $T\left(G(c)\right)$, in other words, is a conjunction of tuples or logical propositions that, in the scene graph, represent the semantic relations. Unlike CIDEr, SPICE can be applied to both small and large datasets because it does not rely on cross-dataset statistics, such as corpus word frequencies. Based on the formulation above, the primary advantage of SPICE is its ability to understand colors and count, as it mainly focuses on objects, attributes, and their association. Its limitation, however, is that it could be ``gamed'' if the candidate captions only include objects, attributes, and their association, and ignore syntax and grammar at the same time. In addition, the aforementioned $n$-gram-based metrics overlook fluency, and SPICE is no exception; therefore, the captions must be well-formed and grammatically correct.%

\subsubsection{\textsc{BERTScore}}%
Zhang et al. \cite{2020_Zhang_BERTScore} introduced \textsc{BERTScore}, an automated evaluation metric for the text generation task. For each token in the candidate sentence, it calculates a score using contextual embeddings, thereby measuring the similarity between that token and each token in the reference sentence. More specifically, \textsc{BERTScore} is based on pre-trained BERT contextual embeddings \cite{2019_Devlin_BERT}, and the similarity score is the summation result of cosine similarities between the token embeddings of candidate and reference sentences. Let $x = \langle x_{1}, \dots, x_{k} \rangle$ be a tokenized reference sentence and $\hat{x} = \langle \hat{x}_{1}, \dots, \hat{x}_{l} \rangle$ be a tokenized candidate sentence. Using contextual embeddings, the elements of both sentences are represented as tokens. In other words, two sequences of vectors, i.e., $\langle \mathbf{x}_{1}, \dots, \mathbf{x}_{k} \rangle$ and $\langle \hat{\mathbf{x}}_{1}, \dots, \hat{\mathbf{x}}_{l} \rangle$, are generated by the embedding model for each tokenized sentence. Assuming two sequences of vectors are pre-normalized, the cosine similarity measure of reference token $x_{i}$ and candidate token $\hat{x}_{j}$ is $\mathbf{x}_{i}^{T} \mathbf{\hat{x}}_{j}$. Now, precision, recall, and F1-measure are calculated as:%
\begin{equation}
    P_{\text{BERT}} = \frac{1}{\lvert \hat{x} \rvert} \sum_{\hat{x}_{j} \in \hat{x}} \max_{x_{i} \in x}\left(\mathbf{x}_{i}^{T} \mathbf{\hat{x}}_{j}\right)\text{,~} R_{\text{BERT}} = \frac{1}{\lvert x \rvert} \sum_{x_{i} \in x} \max_{\hat{x}_{j} \in \hat{x}}\left(\mathbf{x}_{i}^{T} \mathbf{\hat{x}}_{j}\right)\text{,~} F_{\text{BERT}} = \frac{2 \cdot P_{\text{BERT}} \cdot R_{\text{BERT}}}{P_{\textit{BERT}} + R_{\text{BERT}}}\text{.}%
\end{equation}%
As mentioned for CIDEr, less common words can be more informative to indicate sentence similarity \cite{2015_Vedantam_CIDEr}; therefore, to integrate importance weighting, the IDF score of a wordpiece token $w$ is calculated as:%
\begin{equation}
    idf(w) = - \log\left(\frac{1}{M} \sum_{i=1}^{M} \mathbb{I}\left[w \in x^{(i)}\right]\right)\text{,}%
\end{equation}%
where $M$ reference sentences $\{x^{(i)}\}_{i=1}^{M}$ are provided and $\mathbb{I}[\cdot]$ plays the role of indicator function. The complete TF-IDF measure is not calculated for \textsc{BERTScore} because single sentences are being processed, making the TF probably equal to $1$. For instance, recall with importance weighting is calculated as:%
\begin{equation}
    R_{\text{BERT}} = \frac{\sum_{x_{i} \in x} idf\left(x_{i}\right) \cdot \max_{\hat{x}_{j} \in \hat{x}}\left(\mathbf{x}_{i}^{T} \mathbf{\hat{x}}_{j}\right)}{\sum_{x_{i} \in x} idf\left(x_{i}\right)}\text{.}%
\end{equation}%
Pre-normalized vectors force the scores to have the same numerical range as the cosine similarity score (i.e., $[-1, 1]$). To limit the numerical range to $[0, 1]$, the authors of \cite{2020_Zhang_BERTScore} rescale the score based on its empirical lower bound $b$, calculated through Common Crawl monolingual datasets. Hence, the modified versions of precision and recall are as:%
\begin{equation}
    \hat{P}_{\text{BERT}} = \frac{P_{\text{BERT}} - b}{1 - b}\text{,~} \hat{R}_{\text{BERT}} = \frac{R_{\text{BERT}} - b}{1 - b}\text{.}%
\end{equation}%

\subsubsection{\textsc{CLIPScore} and \textsc{RefCLIPScore}}%
Hessel et al. \cite{2021_Hessel_CLIPScore} introduced \textsc{CLIPScore} (CLIP-S) to eliminate the need for a predefined set of reference captions while evaluating generated captions. \textsc{CLIPScore}, as a reference-free metric, is based on CLIP \cite{2021_Radford_CLIP}, a multimodal model that is pre-trained on 400 million image-caption pairings from online sources, allowing for reference-free caption evaluation. To evaluate the quality of a generated caption, CLIP employs two feature extractors: a ViT \cite{2021_Dosovitskiy_ViT} as an image encoder (specifically, the ViT-B/32 variant of ViT is used in \cite{2021_Hessel_CLIPScore}) and a Transformer \cite{2017_Vaswani_Transformer} with masked self-attention as caption/text encoder. After extracting the features, \textsc{CLIPScore} calculates the cosine similarity of generated embeddings. The score is then rescaled to limit the range to $[0, 1]$ rather than the same numerical range as the cosine similarity score (i.e., $[-1, 1]$). Given image visual embedding $\mathbf{v}$ and caption textual embedding $\mathbf{c}$, \textsc{CLIPScore} is calculated as follows:%
\begin{equation}
    \text{CLIP-S}\left(\mathbf{c}, \mathbf{v}\right) = w \cdot \max\left(\cos\left(\mathbf{c}, \mathbf{v}\right), 0\right)\text{,}%
\end{equation}%
where $w$ is a scaling factor set to $2.5$, which stretches the range of cosine similarity values from $[0, 0.4]$ to $[0, 1]$. Hessel et al. \cite{2021_Hessel_CLIPScore} also introduced a reference-augmented variant of \textsc{CLIPScore}, termed \textsc{RefCLIPScore}, that obtains a higher correlation with human judgment. The procedure for calculating \textsc{RefCLIPScore} is like \textsc{CLIPScore} except that the references are also passed through the caption/text encoder of CLIP to generate a vector representation $\mathbf{R}$ for all references. Now, \textsc{RefCLIPScore} is calculated as follows:%
\begin{equation}
    \text{RefCLIP-S}(\mathbf{c}, \mathbf{R}, \mathbf{v}) = \text{H-Mean}\left(\text{CLIP-S}\left(\mathbf{c}, \mathbf{v}\right), \max\left(\max_{\mathbf{r} \in \mathbf{R}} \cos(\mathbf{c}, \mathbf{r}), 0\right)\right)\text{,}
\end{equation}%
where $\text{H-Mean}(\cdot)$ is the harmonic mean of \textsc{CLIPScore} and the cosine similarity of the maximal reference.%

\subsection{Experimental Design}%
A set of preliminary experiments is designed to evaluate the proposed hypotheses, assess the effectiveness of the developed dataset, and examine the influence of model complexity on description generation for traffic scene understanding. These experiments systematically explore the effects of various encoder-decoder combinations, as well as the impact of pre-training on the Flickr8k dataset \cite{2013_Hodosh_Flickr8k}. Rather than presenting a finalized architecture followed by an ablation study, this work adopts a step-by-step experimental approach that culminates in the final model configuration. The experimental steps are primarily structured to identify the optimal encoder-decoder combination that effectively performs with a limited amount of data.%

\subsubsection{Training Configuration}%
\label{subsubsec:TC}
The training configuration is carefully designed to ensure optimal performance and stability. The training process utilizes a batch size of 32 over 100 epochs, with the parameters of the encoder frozen for the first 10 epochs to enable proper initialization of the decoder. This is because all encoders are initialized with pre-trained weights. Two learning rates of 0.00001 (for the encoder) and 0.0005 (for the Intermediate Bridge and the decoder) are set to apply discriminative learning. The learning rates are dynamically adjusted using a StepLR scheduler with a step size of 20 epochs and a decay factor ($\gamma$) of 0.95. The AdamW optimizer is employed for its robust performance, with a weight decay of 0.0001 to mitigate overfitting. The CrossEntropyLoss() function is used as the objective function to measure the discrepancy between the predicted and ground-truth distributions, while ignoring the ``{\textless PAD\textgreater}'' tokens, ensuring effective optimization during training. Gradients are clipped to a maximum norm of 5.0 to ensure stability during backpropagation. Additionally, the vocabulary is filtered using a frequency threshold of 5 to exclude infrequent tokens.%

To address the limited availability of data, the entire dataset comprising 600 images and their corresponding human-generated reference descriptions is utilized during training. The training process incorporates the ``Teacher Forcing'' algorithm, a method specifically designed to optimize the weights of RNNs. This approach involves feeding ground-truth sequence values back into the RNN at each step, thereby constraining the model to remain aligned with the ground-truth sequence throughout the training process.%

\subsubsection{Experimentation Process}%
The present study considers five encoders and two decoders, resulting in ten possible combinations for the encoder-decoder architecture. The proposed hypotheses are applied to these combinations at various levels to determine if the generated descriptions exhibit any improvement. Additionally, every combination is trained once with and once without pre-training on the Flickr8k dataset \cite{2013_Hodosh_Flickr8k} to assess the effect of pre-training on a generic image captioning dataset prior to training on the target dataset. Finally, the output of each model subsequently undergoes a greedy search decoding strategy during inference to generate scene descriptions by selecting the most probable word at each step of the sequence generation process.%

A total of 15 images are selected to assess the performance of the models. These images are divided into three distinct categories, as outlined below:%
\begin{enumerate}
    \item \textbf{\textit{seen\_bdd\_00x}:} Five images from the developed dataset (i.e., 10K Images dataset \cite{2020_Yu_BDD100K} with human-generated reference scene descriptions) that have been utilized during the training process are selected to evaluate whether the models effectively learned from the image-text pairs on which they have been trained. Table~\ref{tab:developed-dataset-p01} and Table~\ref{tab:developed-dataset-p02} present these images with their corresponding reference scene descriptions.%
    \item \textbf{\textit{unseen\_bdd\_00x}:} Five images from the 10K Images dataset \cite{2020_Yu_BDD100K} that have been neither utilized during training nor annotated by human annotators (i.e., the images do not have human-generated reference scene descriptions) are selected to evaluate how the models generalize to unseen data. Table~\ref{tab:dataset-10kimages} presents these images.%
    \item \textbf{\textit{flickr8k\_00x}:} Five images containing various driving scene-related entities are selected from the Flickr8k dataset \cite{2013_Hodosh_Flickr8k} to assess whether the models can predict any such entities present in these images. This evaluation can be considered a step beyond generalization. Table~\ref{tab:dataset-flickr8k} presents these images, excluding their corresponding reference captions, since they are not related to traffic scene understanding.%
\end{enumerate}%

\begin{table}[!t]
    \centering%
    \begin{tabular}{c}
        \toprule%
        \makecell[c]{\textit{unseen\_bdd\_001}\\{\includegraphics[height=1.00in]{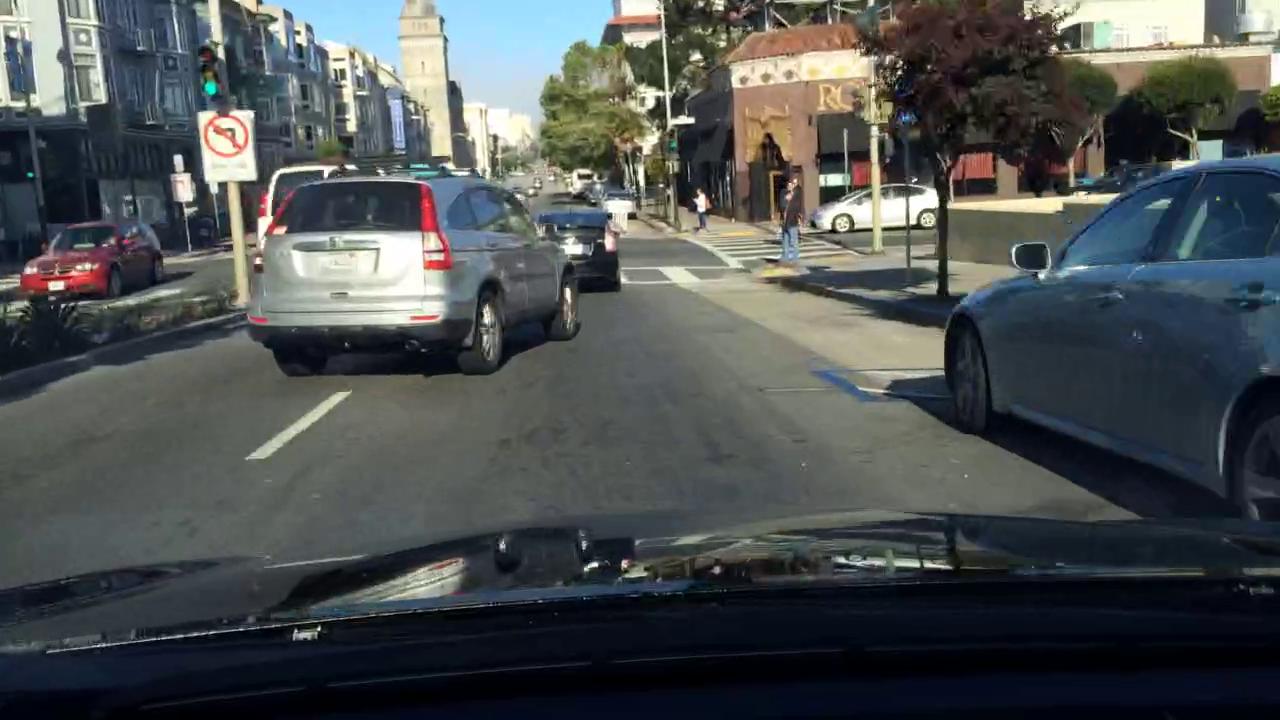}}} \hfill \makecell[c]{\textit{unseen\_bdd\_002}\\{\includegraphics[height=1.00in]{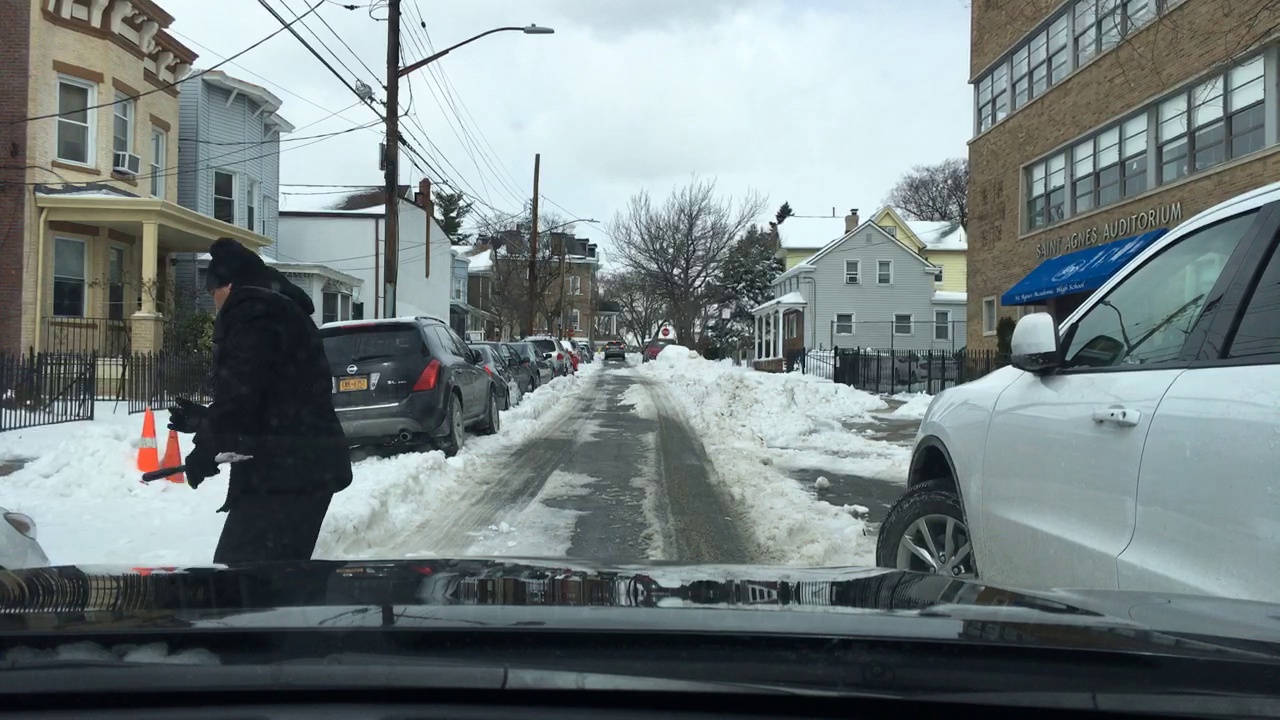}}} \hfill \makecell[c]{\textit{unseen\_bdd\_003}\\{\includegraphics[height=1.00in]{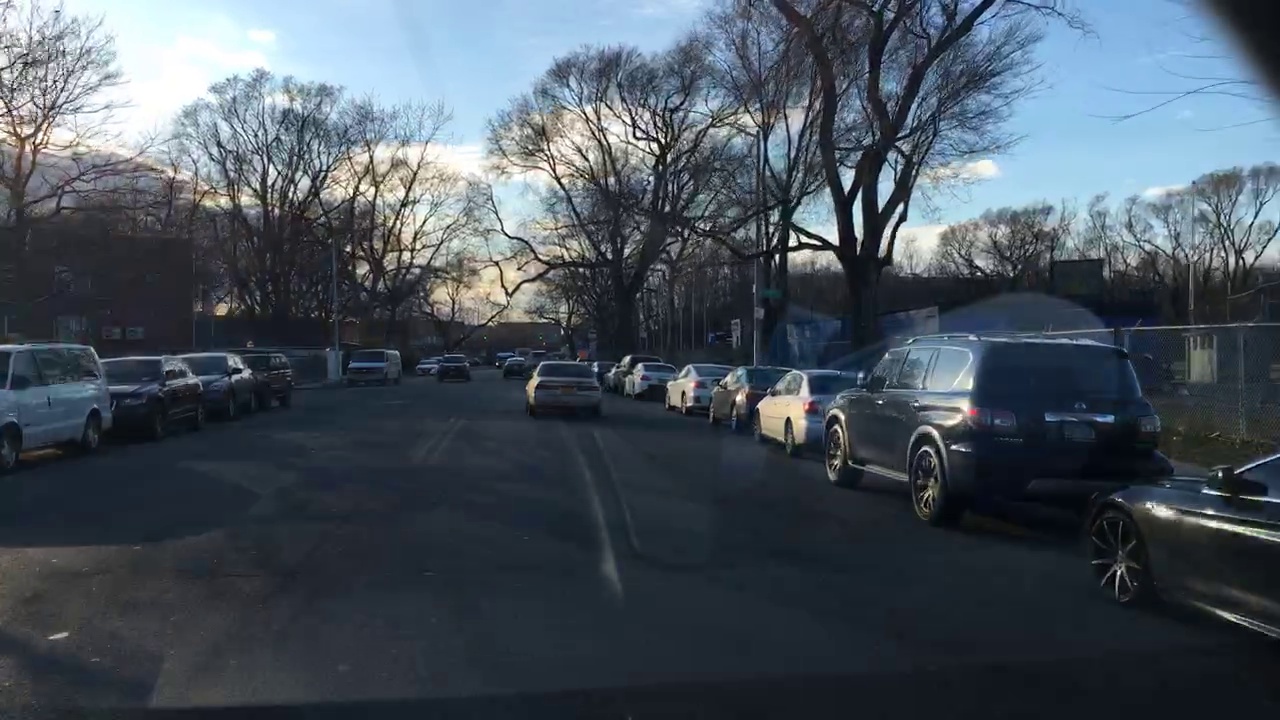}}}\\
        \midrule%
        \makecell[c]{\textit{unseen\_bdd\_004}\\{\includegraphics[height=1.00in]{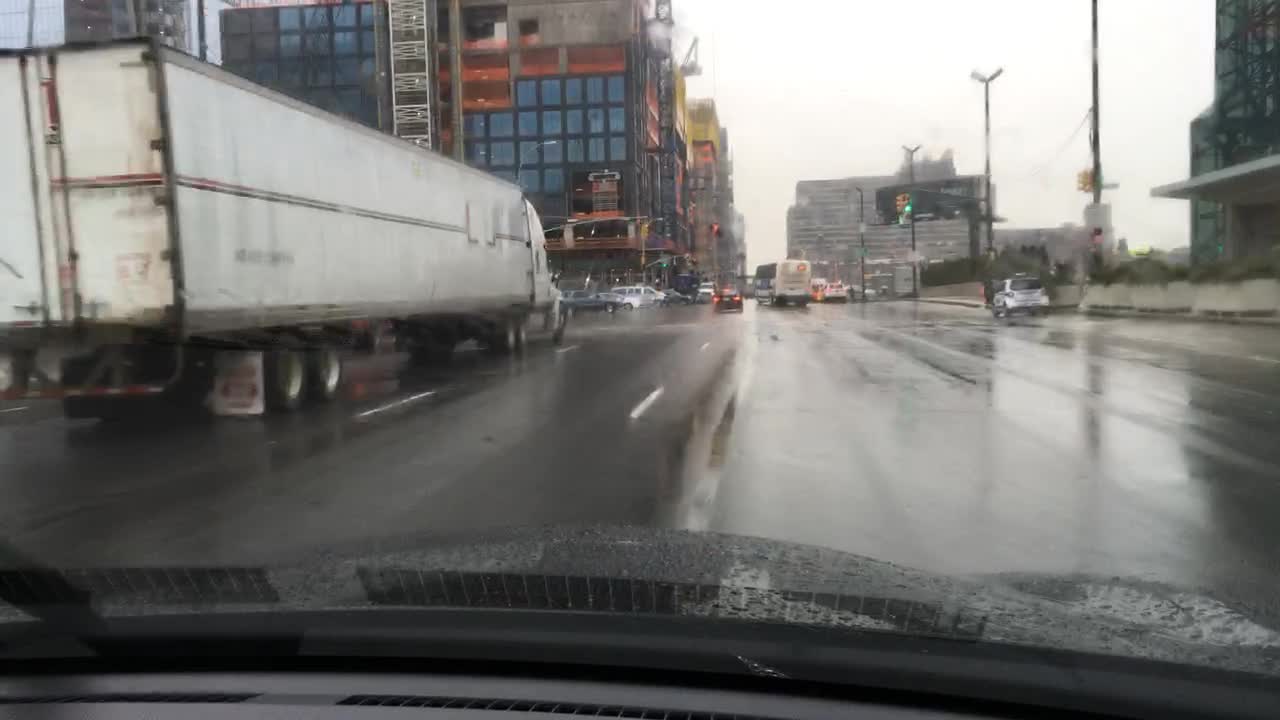}}} \hfill \makecell[c]{\textit{unseen\_bdd\_005}\\{\includegraphics[height=1.00in]{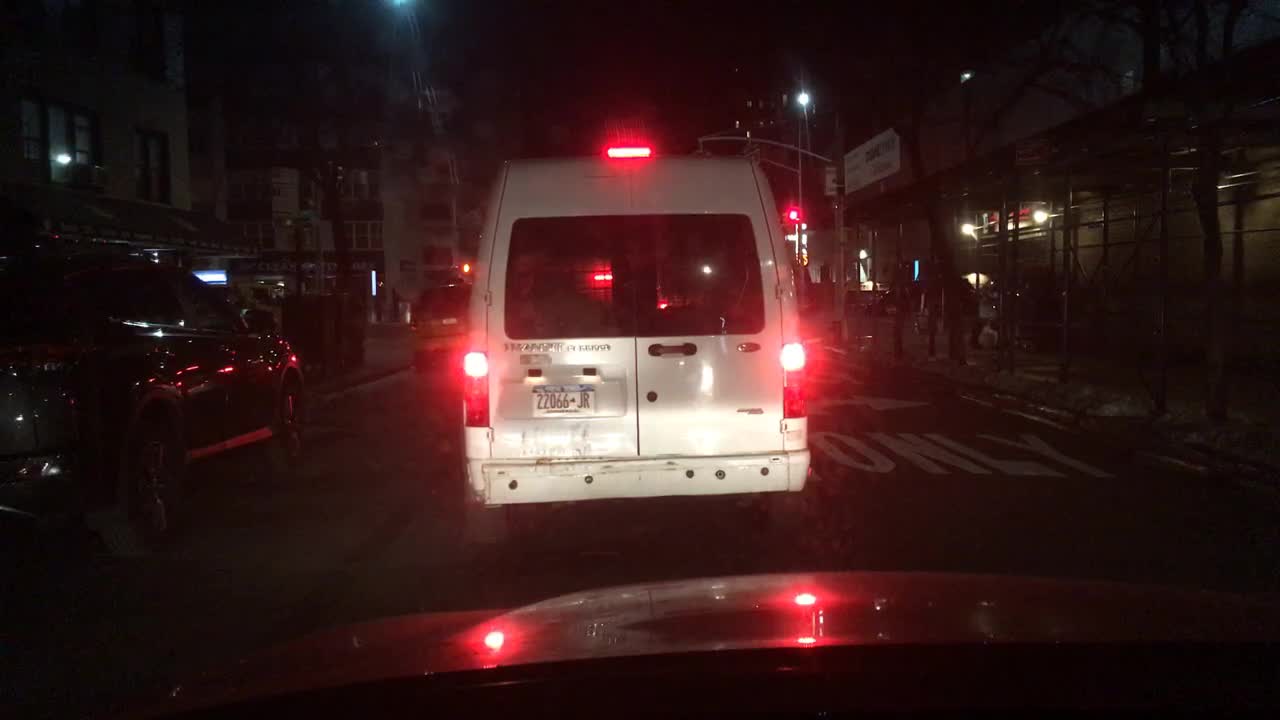}}}\\
        \bottomrule%
    \end{tabular}%
    \caption{Sample Images from 10K Images Dataset \cite{2020_Yu_BDD100K} (Neither Utilized During Training nor Annotated by Human Annotators)}\label{tab:dataset-10kimages}%
\end{table}%

\begin{table}[!t]
    \centering%
    \begin{tabular}{c}
        \toprule%
        \makecell[c]{\textit{flickr8k\_001}\\{\includegraphics[height=1.00in]{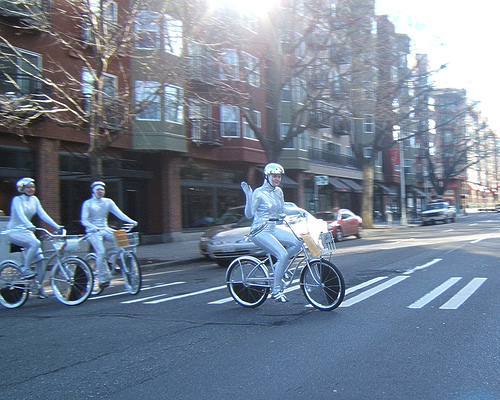}}} \hfill \makecell[c]{\textit{flickr8k\_002}\\{\includegraphics[height=1.00in]{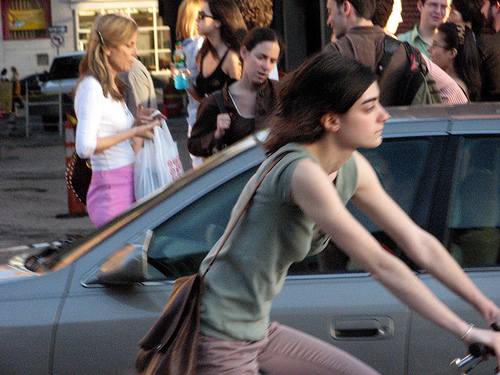}}} \hfill \makecell[c]{\textit{flickr8k\_003}\\{\includegraphics[height=1.00in]{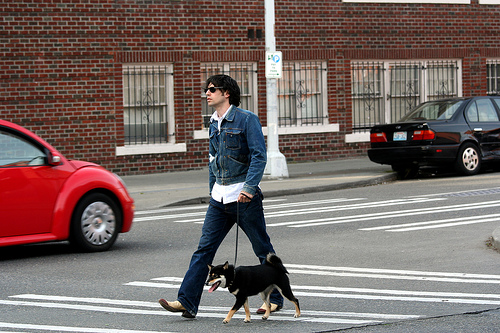}}} \hfill \makecell[c]{\textit{flickr8k\_004}\\{\includegraphics[height=1.00in]{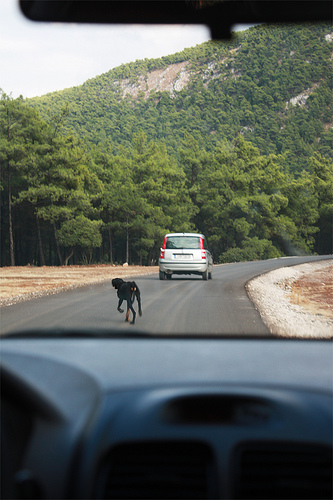}}} \hfill \makecell[c]{\textit{flickr8k\_005}\\{\includegraphics[height=1.00in]{imgs/flickr8k_005.jpg}}}\\
        \bottomrule%
    \end{tabular}%
    \caption{Sample Images from Flickr8k Dataset \cite{2013_Hodosh_Flickr8k}}\label{tab:dataset-flickr8k}%
\end{table}%

Therefore, models will be evaluated first on the \textit{seen\_bdd\_00x} images; after successfully passing the evaluation process, they will be evaluated on the \textit{unseen\_bdd\_00x} images. The \textit{flickr8k\_00x} images will be used for evaluation after observing satisfactory results on \textit{unseen\_bdd\_00x} images.

\section{Analysis and Discussion}%
\label{sec:s04}%
\subsection{Impact of Pre-Training on Flickr8k Dataset}%
Pre-training on the Flickr8k dataset \cite{2013_Hodosh_Flickr8k} did not improve the performance of the models and sometimes even worsened it. This outcome can primarily be attributed to differences in the vocabulary and vocabulary size between the two datasets. With a frequency threshold of 5 to exclude infrequent tokens, the vocabulary size for the Flickr8k dataset \cite{2013_Hodosh_Flickr8k} is 2,992, while for the developed dataset, it is 298, which requires re-initialization of the embedding and prediction layers.%

\subsection{Comparative Analysis of Different Combinations}%
An extensive comparison between different encoder-decoder combinations is conducted in three stages. Each of these stages compares various aspects of the combinations and applies architectural modifications based on the proposed hypotheses to improve the generated scene descriptions.%

\subsubsection{Stage 1 - Analysis for Memory Enhancement and Feature Preservation}%
In this stage, the performance of VGG-16, ResNet-50, DETR, and RT-DETR, when paired with LSTM and xLSTM, is compared among themselves. The initial intention behind this stage is to select a suitable decoder from the two choices of LSTM and xLSTM, as well as to select suitable evaluation metrics that are appropriate for traffic scene understanding through natural language text generation within the context of driving. Table~\ref{tab:s1-seen} and Table~\ref{tab:s1-seen-2} present the results of testing the encoder-decoder combinations on \textit{seen\_bdd\_00x} images shown in Table~\ref{tab:developed-dataset-p01} and Table~\ref{tab:developed-dataset-p02}. These results are followed by Table~\ref{tab:s1-seen-eval} depicting the metric scores for each generated scene description.%

\clearpage%
\newpage%
\newgeometry{left=0.2in, right=0.2in, top=0.2in, bottom=0.2in}%
\begin{table}[!t]
    \centering
    \resizebox{1.00\textwidth}{!}{%
    \begin{tabularx}{\textwidth}{lll}
        \toprule%
        \textbf{} & \textbf{LSTM} & \textbf{xLSTM}\\%
        \midrule%
        \makecell[l]{\rotatebox{90}{\textbf{VGG-16}}} & \makecell[X]{\scriptsize{\texttt{\fbox{001:} it is clear daytime , many cars are on the right , many cars are parked on both sides of the street , and a crosswalk is ahead nearby .}}\\ \scriptsize{\texttt{\fbox{002:} it is clear nighttime , it is a two - way street , one black suv is braking in front with its brake lights on , the ego lane is the leftmost lane , and 1 black car is braking in front in the ego lane with its brake lights on .}}\\ \scriptsize{\texttt{\fbox{003:} a yellow taxi is driving in the ego lane with its left turn signal flashing , the ego lane is the rightmost lane , and a car is driving in the left lane ahead .}}\\ \scriptsize{\texttt{\fbox{004:} it is a two - lane street .}}\\ \scriptsize{\texttt{\fbox{005:} it is raining , the street is wet and slippery because of the rain , and many vehicles are parked on both sides of the street .}}} & \makecell[X]{\scriptsize{\texttt{\fbox{001:} it is clear daytime .}}\\ \scriptsize{\texttt{\fbox{002:} it is clear nighttime , it is a two - way street , and 1 black suv is braking in front with its brake lights on .}}\\ \scriptsize{\texttt{\fbox{003:} a [ school zone ] sign is on the right side of the street .}}\\ \scriptsize{\texttt{\fbox{004:} it is clear daytime , there is traffic congestion ahead , many vehicles are braking with their brake lights on , 1 black car with its brake lights on is braking in the ego lane in front , 1 blue van is braking with its brake lights on in front is braking with its brake lights on in front nearby , and the traffic light is red .}}\\ \scriptsize{\texttt{\fbox{005:} it is a one - way , narrow street , and the street is wet and slippery because of the rain .}}}\\%
        \midrule%
        \makecell[l]{\rotatebox{90}{\textbf{ResNet-50}}} & \makecell[X]{\scriptsize{\texttt{\fbox{001:} it is a residential area , the traffic light is green , and 1 white car and 1 black car are driving in the ego lane nearby .}}\\ \scriptsize{\texttt{\fbox{002:} it is a two - way street , and 1 black suv is driving in front in the ego lane straight ahead .}}\\ \scriptsize{\texttt{\fbox{003:} a bus is braking in the right lane with its brake lights on , and 1 person is walking on the sidewalk on the right .}}\\ \scriptsize{\texttt{\fbox{004:} it is clear daytime , there is traffic congestion ahead , many vehicles are braking with their brake lights on , the traffic light is red at intersection , 1 black car with its brake lights on is braking in the ego lane in front at a short distance away , and there is a [ direction ] sign with green background on the left side of the street .}}\\ \scriptsize{\texttt{\fbox{005:} it is raining , and the road is wet and slippery because of the rain .}}} & \makecell[X]{\scriptsize{\texttt{\fbox{001:} it is a multi - lane street .}}\\ \scriptsize{\texttt{\fbox{002:} it is clear nighttime , it is a two - way street , one black suv is braking in front with its brake lights on , the ego lane is the leftmost lane , an intersection is ahead , the traffic light is green at the intersection , one silver car is driving in the ego lane , the street has multiple lanes , there is a white ahead , the right lane is clear , and several vehicles are parked on both sides of the street some distance ahead .}}\\ \scriptsize{\texttt{\fbox{003:} a pedestrian is on the right side of the road .}}\\ \scriptsize{\texttt{\fbox{004:} it is clear daytime , it is a two - way street , the traffic light is green , a [ one way ] sign pointing straight , many vehicles are parked on both sides of the street , the road has 2 lanes , the ego lane is the leftmost lane , 1 car is driving in the ego lane in front , there is a crosswalk in front , the traffic lights are green , many vehicles are parked on both sides of the street , one [ on the street , and there is a [ no left turn ] sign on the left side of the street some distance ahead , there are 2 [ on the right side of the street some distance ahead , and the traffic light is green .}}\\ \scriptsize{\texttt{\fbox{005:} it is a one - way , narrow street , and the street is wet and slippery because of the rain .}}}\\%
        \bottomrule%
    \end{tabularx}%
    }%
    \caption{Generated Scene Descriptions in Stage 1 Based on \textit{seen\_bdd\_00x} Images - Part 1}\label{tab:s1-seen}%
\end{table}%
\thispagestyle{empty}%
\restoregeometry
\newpage%
\clearpage
\newpage%

\clearpage%
\newpage%
\newgeometry{left=0.2in, right=0.2in, top=0.2in, bottom=0.2in}%
\begin{table}[!t]
    \centering
    \resizebox{1.00\textwidth}{!}{%
    \begin{tabularx}{\textwidth}{lll}
        \toprule%
        \textbf{} & \textbf{LSTM} & \textbf{xLSTM}\\%
        \midrule%
        \makecell[l]{\rotatebox{90}{\textbf{DETR}}} & \makecell[X]{\scriptsize{\texttt{\fbox{001:} it is a residential area .}}\\ \scriptsize{\texttt{\fbox{002:} it is clear nighttime , it is an intersection , and 1 black suv is braking in the ego lane in front with its brake lights on .}}\\ \scriptsize{\texttt{\fbox{003:} it is clear daytime , and the ego lane is the leftmost lane .}}\\ \scriptsize{\texttt{\fbox{004:} it is traffic congestion .}}\\ \scriptsize{\texttt{\fbox{005:} it is raining , and the road is wet and slippery because of the rain .}}} & \makecell[X]{\scriptsize{\texttt{\fbox{001:} it is a multi - lane street .}}\\ \scriptsize{\texttt{\fbox{002:} it is clear nighttime , it is a two - way street , one black suv is braking in front with its brake lights on , many vehicles is parked on the right side of the street , and there is a [ school zone ] sign on the right side of the street .}}\\ \scriptsize{\texttt{\fbox{003:} a pedestrian is on the right side of there is crossing the right sidewalk .}}\\ \scriptsize{\texttt{\fbox{004:} it is clear daytime .}}\\ \scriptsize{\texttt{\fbox{005:} it is daytime , it is raining , it is a one - way , narrow street , the road is wet and slippery because of the rain , many vehicles are parked on both sides of the street , and a vehicle is braking in the ego lane some distance ahead with its taillights and brake lights on .}}}\\%
        \midrule%
        \makecell[l]{\rotatebox{90}{\textbf{RT-DETR}}} & \makecell[X]{\scriptsize{\texttt{\fbox{001:} the ego lane is the rightmost lane .}}\\ \scriptsize{\texttt{\fbox{002:} the ego lane is the rightmost lane .}}\\ \scriptsize{\texttt{\fbox{003:} the ego lane is the rightmost lane .}}\\ \scriptsize{\texttt{\fbox{004:} the ego lane is the rightmost lane .}}\\ \scriptsize{\texttt{\fbox{005:} the ego lane is the rightmost lane .}}} & \makecell[X]{\scriptsize{\texttt{\fbox{001:} it is a narrow , one - way street , 1 white truck is parked in front of at on because brake lights on the left side , 1 pedestrian is on the opposite lane , and 1 pedestrian is on the right sidewalk .}}\\ \scriptsize{\texttt{\fbox{002:} it is a narrow one black cliff pedestrian an unpaved ] signs are in front , and there are in front of the and there are in front .}}\\ \scriptsize{\texttt{\fbox{003:} it is an intersection .}}\\ \scriptsize{\texttt{\fbox{004:} it is an intersection .}}\\ \scriptsize{\texttt{\fbox{005:} it is an intersection .}}}\\%
        \bottomrule%
    \end{tabularx}%
    }%
    \caption{Generated Scene Descriptions in Stage 1 Based on \textit{seen\_bdd\_00x} Images - Part 2}\label{tab:s1-seen-2}%
\end{table}%
\thispagestyle{empty}%
\restoregeometry
\newpage%
\clearpage
\newpage%

\begin{table}[!tb]
    \centering
    \resizebox{1.00\textwidth}{!}{%
    \begin{tabular}{lllllllllllll}
        \toprule%
        \textbf{} & \textbf{id} & \makecell[l]{\textbf{BLEU-1}\\ LSTM $\mid$ xLSTM} & \makecell[l]{\textbf{BLEU-2}\\ LSTM $\mid$ xLSTM} & \makecell[l]{\textbf{BLEU-3}\\ LSTM $\mid$ xLSTM} & \makecell[l]{\textbf{BLEU-4}\\ LSTM $\mid$ xLSTM} & \makecell[l]{\textbf{ROUGE-L}\\ LSTM $\mid$ xLSTM} & \makecell[l]{\textbf{METEOR}\\ LSTM $\mid$ xLSTM} & \makecell[l]{\textbf{CIDEr}\\ LSTM $\mid$ xLSTM} & \makecell[l]{\textbf{SPICE}\\ LSTM $\mid$ xLSTM} & \makecell[l]{\textbf{\textsc{BERTScore}}\\ LSTM $\mid$ xLSTM} & \makecell[l]{\textbf{\textsc{CLIPScore}}\\ LSTM $\mid$ xLSTM} & \makecell[l]{\textbf{\textsc{RefCLIPScore}}\\ LSTM $\mid$ xLSTM}\\%
        \midrule%
        \multirow{5}{*}{\rotatebox{90}{\textbf{VGG-16}}} & 001 & {0.8000} $\mid$ {1.0000} & {0.6848} $\mid$ {1.0000} & {0.5858} $\mid$ {1.0000} & {0.4940} $\mid$ {1.0000} & {0.8082} $\mid$ {1.0000} & {0.2984} $\mid$ {1.0000} & {0.0268} $\mid$ {1.0000} & {0.2927} $\mid$ {0.1250} & {0.9144} $\mid$ {1.0000} & {0.2817} $\mid$ {0.2150} & {0.4096} $\mid$ {0.3540}\\%
        & 002 & {0.7500} $\mid$ {1.0000} & {0.7376} $\mid$ {1.0000} & {0.7317} $\mid$ {1.0000} & {0.7273} $\mid$ {1.0000} & {0.8688} $\mid$ {1.0000} & {0.5134} $\mid$ {1.0000} & {0.4989} $\mid$ {1.5706} & {0.6522} $\mid$ {0.5263} & {0.9433} $\mid$ {1.0000} & {0.3092} $\mid$ {0.3014} & {0.4276} $\mid$ {0.4423}\\%
        & 003 & {0.7429} $\mid$ {1.0000} & {0.6610} $\mid$ {0.9636} & {0.5416} $\mid$ {0.9499} & {0.4598} $\mid$ {0.9415} & {0.7069} $\mid$ {0.9621} & {0.3925} $\mid$ {0.5535} & {0.0011} $\mid$ {1.3382} & {0.2759} $\mid$ {0.3774} & {0.9057} $\mid$ {0.9881} & {0.2663} $\mid$ {0.2605} & {N/A} $\mid$ {N/A}\\%
        & 004 & {1.0000} $\mid$ {0.7109} & {0.8452} $\mid$ {0.6631} & {0.7095} $\mid$ {0.6178} & {0.6148} $\mid$ {0.5754} & {0.8750} $\mid$ {0.8541} & {0.3997} $\mid$ {0.3896} & {0.3056} $\mid$ {0.1184} & {0.0476} $\mid$ {0.5161} & {0.8939} $\mid$ {0.9571} & {0.2807} $\mid$ {0.3021} & {N/A} $\mid$ {N/A}\\%
        & 005 & {0.8623} $\mid$ {1.0000} & {0.8623} $\mid$ {1.0000} & {0.8506} $\mid$ {1.0000} & {0.8352} $\mid$ {1.0000} & {0.9845} $\mid$ {1.0000} & {0.5315} $\mid$ {1.0000} & {1.1649} $\mid$ {1.7802} & {0.4848} $\mid$ {0.4375} & {0.9604} $\mid$ {1.0000} & {0.3139} $\mid$ {0.2653} & {0.4566} $\mid$ {0.4039}\\%
        \midrule%
        \multirow{5}{*}{\rotatebox{90}{\textbf{ResNet-50}}} & 001 & {0.7500} $\mid$ {1.0000} & {0.6236} $\mid$ {1.0000} & {0.5480} $\mid$ {1.0000} & {0.4790} $\mid$ {1.0000} & {0.7379} $\mid$ {1.0000} & {0.4151} $\mid$ {1.0000} & {0.0831} $\mid$ {1.0000} & {0.3415} $\mid$ {0.1818} & {0.9283} $\mid$ {1.0000} & {0.2632} $\mid$ {0.2846} & {0.3780} $\mid$ {0.4245}\\%
        & 002 & {0.8696} $\mid$ {0.5604} & {0.7952} $\mid$ {0.5467} & {0.7123} $\mid$ {0.5246} & {0.6520} $\mid$ {0.5093} & {0.8978} $\mid$ {0.7315} & {0.3860} $\mid$ {0.5575} & {0.9434} $\mid$ {0.0000} & {0.3784} $\mid$ {0.6207} & {0.9167} $\mid$ {0.9452} & {0.3059} $\mid$ {N/A} & {0.4384} $\mid$ {N/A}\\%
        & 003 & {0.8846} $\mid$ {1.0000} & {0.7756} $\mid$ {1.0000} & {0.7053} $\mid$ {1.0000} & {0.6541} $\mid$ {1.0000} & {0.8905} $\mid$ {1.0000} & {0.5266} $\mid$ {1.0000} & {0.0699} $\mid$ {1.6407} & {0.2545} $\mid$ {0.2449} & {0.9528} $\mid$ {1.0000} & {0.2870} $\mid$ {0.2775} & {N/A} $\mid$ {N/A}\\%
        & 004 & {0.7263} $\mid$ {0.4161} & {0.6572} $\mid$ {0.3410} & {0.5887} $\mid$ {0.2625} & {0.5471} $\mid$ {0.2047} & {0.8506} $\mid$ {0.4364} & {0.3632} $\mid$ {0.2578} & {0.1955} $\mid$ {0.0000} & {0.4737} $\mid$ {0.3077} & {0.9494} $\mid$ {0.8679} & {0.2967} $\mid$ {N/A} & {N/A} $\mid$ {N/A}\\%
        & 005 & {1.0000} $\mid$ {1.0000} & {1.0000} $\mid$ {1.0000} & {1.0000} $\mid$ {1.0000} & {1.0000} $\mid$ {1.0000} & {1.0000} $\mid$ {1.0000} & {1.0000} $\mid$ {1.0000} & {2.0714} $\mid$ {1.7802} & {0.2759} $\mid$ {0.4375} & {1.0000} $\mid$ {1.0000} & {0.2806} $\mid$ {0.2653} & {0.4313} $\mid$ {0.4039}\\%
        \midrule%
        \multirow{5}{*}{\rotatebox{90}{\textbf{DETR}}} & 001 & {1.0000} $\mid$ {1.0000} & {1.0000} $\mid$ {1.0000} & {1.0000} $\mid$ {1.0000} & {1.0000} $\mid$ {1.0000} & {1.0000} $\mid$ {1.0000} & {1.0000} $\mid$ {1.0000} & {1.0000} $\mid$ {1.0000} & {0.1250} $\mid$ {0.1818} & {1.0000} $\mid$ {1.0000} & {0.2070} $\mid$ {0.2847} & {0.3352} $\mid$ {0.4245}\\%
        & 002 & {0.9643} $\mid$ {0.6727} & {0.9063} $\mid$ {0.6011} & {0.8285} $\mid$ {0.5617} & {0.7512} $\mid$ {0.5291} & {0.8995} $\mid$ {0.7854} & {0.5157} $\mid$ {0.4959} & {1.0645} $\mid$ {0.2148} & {0.5366} $\mid$ {0.4400} & {0.9600} $\mid$ {0.9409} & {0.2916} $\mid$ {0.2858} & {0.4269} $\mid$ {0.4152}\\%
        & 003 & {1.0000} $\mid$ {0.8667} & {0.9608} $\mid$ {0.7464} & {0.9163} $\mid$ {0.6359} & {0.8365} $\mid$ {0.5721} & {0.6784} $\mid$ {0.8912} & {0.3295} $\mid$ {0.4185} & {0.4257} $\mid$ {0.9089} & {0.2449} $\mid$ {0.1600} & {0.9144} $\mid$ {0.9467} & {0.2251} $\mid$ {0.2726} & {N/A} $\mid$ {N/A}\\%
        & 004 & {1.0000} $\mid$ {1.0000} & {0.8660} $\mid$ {1.0000} & {0.6300} $\mid$ {1.0000} & {0.0001} $\mid$ {1.0000} & {0.7722} $\mid$ {1.0000} & {0.3553} $\mid$ {1.0000} & {0.5347} $\mid$ {1.0000} & {0.0952} $\mid$ {0.0952} & {0.9484} $\mid$ {1.0000} & {0.2656} $\mid$ {0.2360} & {N/A} $\mid$ {N/A}\\%
        & 005 & {1.0000} $\mid$ {1.0000} & {1.0000} $\mid$ {1.0000} & {1.0000} $\mid$ {0.9942} & {1.0000} $\mid$ {0.9868} & {1.0000} $\mid$ {0.9931} & {1.0000} $\mid$ {0.6320} & {2.0714} $\mid$ {0.9625} & {0.2759} $\mid$ {0.9130} & {1.0000} $\mid$ {0.9989} & {0.2806} $\mid$ {0.2877} & {0.4313} $\mid$ {0.4250}\\%
        \midrule%
        \multirow{5}{*}{\rotatebox{90}{\textbf{RT-DETR}}} & 001 & {0.8750} $\mid$ {0.5170} & {0.7071} $\mid$ {0.3552} & {0.4368} $\mid$ {0.2359} & {0.0001} $\mid$ {0.1293} & {0.4946} $\mid$ {0.5520} & {0.1879} $\mid$ {0.2158} & {0.0304} $\mid$ {0.0202} & {0.1765} $\mid$ {0.2353} & {0.8839} $\mid$ {0.8824} & {0.2269} $\mid$ {0.3072} & {0.3549} $\mid$ {0.4115}\\%
        & 002 & {0.8750} $\mid$ {0.4483} & {0.7071} $\mid$ {0.2829} & {0.6300} $\mid$ {0.1437} & {0.5623} $\mid$ {0.0000} & {0.5567} $\mid$ {0.4845} & {0.1612} $\mid$ {0.1666} & {0.0052} $\mid$ {0.0382} & {0.1875} $\mid$ {0.1081} & {0.8920} $\mid$ {0.8255} & {0.2239} $\mid$ {0.2035} & {0.3492} $\mid$ {0.3197}\\%
        & 003 & {0.6014} $\mid$ {0.1807} & {0.4860} $\mid$ {0.1167} & {0.4330} $\mid$ {0.0000} & {0.3865} $\mid$ {0.0000} & {0.5503} $\mid$ {0.2909} & {0.2300} $\mid$ {0.0660} & {0.0403} $\mid$ {0.0000} & {0.1277} $\mid$ {0.0000} & {0.9151} $\mid$ {0.8780} & {0.2008} $\mid$ {0.2657} & {N/A} $\mid$ {N/A}\\%
        & 004 & {0.8750} $\mid$ {0.8000} & {0.6124} $\mid$ {0.4472} & {0.3969} $\mid$ {0.0000} & {0.0001} $\mid$ {0.0000} & {0.5145} $\mid$ {0.6685} & {0.0962} $\mid$ {0.1540} & {0.0123} $\mid$ {0.0033} & {0.0909} $\mid$ {0.0488} & {0.8695} $\mid$ {0.8824} & {0.2334} $\mid$ {0.2602} & {N/A} $\mid$ {N/A}\\%
        & 005 & {0.6619} $\mid$ {0.2696} & {0.4085} $\mid$ {0.1740} & {0.2906} $\mid$ {0.0000} & {0.0000} $\mid$ {0.0000} & {0.4388} $\mid$ {0.4076} & {0.1160} $\mid$ {0.1108} & {0.0000} $\mid$ {0.0000} & {0.1379} $\mid$ {0.0000} & {0.8461} $\mid$ {0.8896} & {0.2102} $\mid$ {0.2579} & {0.3306} $\mid$ {0.3942}\\%
        \bottomrule%
    \end{tabular}%
    }%
    \caption{Evaluation Scores for Generated Scene Descriptions in Stage 1 Based on \textit{seen\_bdd\_00x} Images}\label{tab:s1-seen-eval}%
\end{table}%

Evaluating the quality of the generated descriptions with respect to the reference descriptions in Table~\ref{tab:developed-dataset-p01} and Table~\ref{tab:developed-dataset-p02} via human judgment shows that most of the metrics in Table~\ref{tab:s1-seen-eval} appear unsuitable for evaluating the quality and correctness of the generated descriptions. First of all, as Table~\ref{tab:s1-seen-eval} depicts, \textsc{CLIPScore} and \textsc{RefCLIPScore} produce some {N/A} values, which is due to the context length limitation they have that restricts them from processing generated or reference descriptions exceeding the maximum allowed length for the text encoder of CLIP model. Methods such as chunking and averaging can be leveraged in such situations, but there is no guarantee of the evaluation accuracy. Additionally, lengthy descriptions of the driving scene are necessary when the scene contains numerous details. BLEU is the next unsuitable metric in this context because it outputs 1.0000 when the generated text is any of the reference descriptions; this is unacceptable since giving a score of 1.0000 to ``\texttt{it is clear daytime .}'' is incorrect and misleading in understanding the scene due to lack of detailed scene information. The same reasoning applies to ROUGE-L, METEOR, and \textsc{BERTScore}, making them unsuitable options as well.%

On the other hand, CIDEr and SPICE scores appear well-aligned with human judgment. First, it should be noted that the CIDEr score is in the range of $[0, 10]$, unlike other metrics. Second, when computing the CIDEr score, one cannot provide the metric with only one generated description and its corresponding set of reference descriptions, as CIDEr incorrectly produces 0.0000; this is due to how the IDF is computed in the CIDEr metric implementation that providing few sentences leads to IDF values that result in a CIDEr score of zero. Third, CIDEr is more focused on literal similarity and consensus among reference descriptions (it prioritizes literal accuracy), while SPICE emphasizes semantic meaning and relationships within the image (it prioritizes semantic understanding). Overall, CIDEr may not capture semantic similarity as it focuses on rare words, while SPICE captures semantic meaning and is robust to linguistic variations. Therefore, CIDEr and SPICE appear to be more suitable metrics for this task.%

The next step is to select a suitable decoder from the two options. The predicted textual descriptions demonstrate that the xLSTM, as opposed to the LSTM, predominantly generates longer sentences. This behavior is expected since xLSTM incorporates an improved memory mechanism and offers a greater number of hyperparameters for tuning. It should be acknowledged, however, that LSTM performed comparably in terms of the correctness of the generated sentences. As depicted in Table~\ref{tab:s1-seen} and Table~\ref{tab:s1-seen-2}, when paired with DETR and RT-DETR, xLSTM was able to generate longer sentences and, because of a greater number of hyperparameters, the quality of generated sentences can be improved with a more advanced training strategy and careful hyperparameter tuning. Therefore, in this study, the choice of decoder for the encoder-decoder architecture would be xLSTM.%

An important observation from the generated outputs is that models with RT-DETR as encoder failed to generate acceptable results, while DETR showed a better performance. Specifically, the use of RT-DETR can be considered entirely ineffective, whereas DETR was able to generate some acceptable sentences. Furthermore, the performance of VGG-16 and ResNet-50 surpassed that of DETR, indicating that features extracted from the earlier layers play a more critical role in achieving optimal results. This conclusion is supported by the fact that DETR and RT-DETR employ ResNet-50 as their backbone; yet, ResNet-50 alone outperforms both in describing the scene. In light of this, two new assumptions are made, leading to the conduct of new experiments as Stage 2. These assumptions are as follows:%
\begin{itemize}
    \item \textit{Would applying self-attention and cross-attention mechanisms to the output of the encoder lead to improvement in description generation?}%
    \item \textit{Since RT-DETR has an encoder-decoder architecture, would the output of its encoder, rather than its decoder, lead to improvement in description generation?}%
\end{itemize}%

\subsubsection{Stage 2 - Analysis for Feature Preservation, Intra-Encoder Fusion, and Inter-Encoder Fusion}%
At this stage, the performance of VGG-16, when combined with self-attention and cross-attention mechanisms, is compared with that of RT-DETR, which utilizes only the output of its encoder as input to the decoder. The primary underlying reason for this stage is to investigate whether incorporating an attention mechanism into CNN-based encoders can enhance the predicted results. The second reason is to investigate whether extracting features from the earlier layers of RT-DETR would enhance the predicted sentences. Table~\ref{tab:s2-seen} and Table~\ref{tab:s2-seen-2} present the results of testing the encoder-decoder combinations on \textit{seen\_bdd\_00x} images, where three variations of VGG-16 are compared against three variations of RT-DETR (only the encoder of RT-DETR).%

\clearpage%
\newpage%
\newgeometry{left=0.2in, right=0.2in, top=0.2in, bottom=0.2in}%
\begin{table}[!t]
    \centering
    \begin{tabularx}{\textwidth}{ll}
        \toprule%
        \makecell[l]{\rotatebox{90}{\textbf{VGG-16-E01}}} & \makecell[X]{\scriptsize{\texttt{\fbox{001:} it is clear daytime , it is a multi - lane street , a crosswalk is ahead , one white car is driving in the ego lane nearby , no pedestrians are on the crosswalk , many cars are on the street , and it is a residential area .}}\\ \scriptsize{\texttt{\fbox{002:} it is clear nighttime .}}\\ \scriptsize{\texttt{\fbox{003:} it is a two - lane street , and a crosswalk is ahead in front .}}\\ \scriptsize{\texttt{\fbox{004:} it is clear daytime .}}\\ \scriptsize{\texttt{\fbox{005:} it is a one - way , narrow street , and the street is wet and slippery because of the rain .}}}\\%
        \midrule%
        \makecell[l]{\rotatebox{90}{\textbf{VGG-16-E02}}} & \makecell[X]{\scriptsize{\texttt{\fbox{001:} it is clear daytime , it is a multi - lane street , a crosswalk is ahead , one white car is driving in the ego lane nearby , no pedestrians are on the crosswalk , many cars are on the street , and it is a residential area .}}\\ \scriptsize{\texttt{\fbox{002:} it is clear nighttime , it is a two - way street , one black suv is braking in front with its brake lights on , the ego lane is the leftmost lane , an intersection is ahead , 2 vehicles are in the far ahead , the traffic light is green at the intersection , many vehicles are parked on both sides of the street , there is a short distance away in the opposite direction ] signs with a green for left side street in front some far ahead , and the ego lane is clear .}}\\ \scriptsize{\texttt{\fbox{003:} a pedestrian is on the right side of the road .}}\\ \scriptsize{\texttt{\fbox{004:} it is clear daytime .}}\\ \scriptsize{\texttt{\fbox{005:} it is daytime , and it is raining .}}}\\%
        \midrule%
        \makecell[l]{\rotatebox{90}{\textbf{VGG-ResNet}}} & \makecell[X]{\scriptsize{\texttt{\fbox{001:} it is a multi - lane street .}}\\ \scriptsize{\texttt{\fbox{002:} it is clear nighttime .}}\\ \scriptsize{\texttt{\fbox{003:} a [ school zone ] sign is on the right side of the street ahead .}}\\ \scriptsize{\texttt{\fbox{004:} it is clear daytime .}}\\ \scriptsize{\texttt{\fbox{005:} it is daytime , and it is raining .}}}\\%
        \bottomrule%
    \end{tabularx}%
    \caption{Generated Scene Descriptions in Stage 2 Based on \textit{seen\_bdd\_00x} Images - Part 1}\label{tab:s2-seen}%
\end{table}%
\thispagestyle{empty}%
\restoregeometry
\newpage%
\clearpage
\newpage%

\clearpage%
\newpage%
\newgeometry{left=0.2in, right=0.2in, top=0.2in, bottom=0.2in}%
\begin{table}[!t]
    \centering
    \begin{tabularx}{\textwidth}{ll}
        \toprule%
        \makecell[l]{\rotatebox{90}{\textbf{RT-DETR-E01}}} & \makecell[X]{\scriptsize{\texttt{\fbox{001:} many in the intersection because the crosswalk are in front in the ego lane , and the crosswalk driving many distance on .}}\\ \scriptsize{\texttt{\fbox{002:} the traffic light with green in front of blocking the way keep driving over a distance on front over the street black parked on front .}}\\ \scriptsize{\texttt{\fbox{003:} a crosswalk are of the snow snow ] on the snow both on a parked on front .}}\\ \scriptsize{\texttt{\fbox{004:} many riding on the at ] signs on the at ] signs on the at ] signs on the at ] signs on the at ] signs on the in 2 in 2 in 2 in 2 in 2 in 2 in 2 in 2 in 2 in 2 in 2 in 2 in the ego lane so front .}}\\ \scriptsize{\texttt{\fbox{005:} many <SOS> parked on front front in front both on front in 2 far on front .}}}\\%
        \midrule%
        \makecell[l]{\rotatebox{90}{\textbf{RT-DETR-E02}}} & \makecell[X]{\scriptsize{\texttt{\fbox{001:} it is a residential area , a yellow yellow yellow yellow yellow yellow yellow yellow yellow yellow yellow yellow yellow yellow yellow yellow yellow yellow yellow yellow yellow yellow yellow yellow yellow yellow yellow yellow yellow yellow yellow yellow yellow yellow yellow yellow yellow yellow yellow yellow yellow yellow yellow yellow yellow yellow yellow yellow yellow yellow yellow yellow yellow yellow yellow yellow lines in a bus is clear and 1 car is driving are driving in the street are green the road .}}\\ \scriptsize{\texttt{\fbox{002:} it is clear daytime , one yellow yellow yellow yellow yellow yellow yellow yellow yellow yellow yellow yellow yellow yellow yellow yellow yellow yellow yellow yellow yellow yellow yellow yellow yellow yellow yellow bus is some at the ego at a yellow bus is driving at the right standing of the ego at the ego intersection is clear and and a person is on the right side of the street are green the road .}}\\ \scriptsize{\texttt{\fbox{003:} it is a two - bus is braking in the ego lane some bus is driving in the street with a yellow bus is driving , and a yellow bus is driving on the street with its brake lights on the left and a bus is driving of the street are on the left and a <UNK> are driving of the traffic congestion .}}\\ \scriptsize{\texttt{\fbox{004:}  it is clear daytime , a one yellow yellow yellow yellow yellow yellow yellow yellow yellow yellow yellow yellow yellow yellow bus is in the right lane the right side a traffic the traffic lights ahead are nearby , and 1 traffic the ego standing some distance ahead }}\\ \scriptsize{\texttt{\fbox{005:} it is a residential , and a car is driving .}}}\\%
        \midrule%
        \makecell[l]{\rotatebox{90}{\textbf{RT-DETR-E03}}} & \makecell[X]{\scriptsize{\texttt{\fbox{001:} . , are because because because because because because because because . . . . . . . . . . . . . . . .}}\\ \scriptsize{\texttt{\fbox{002:} . car some because . . . . . . . . . . . . . . . . . . . . . . . . . . . . . . . . . . . . . . . . . . . . . . . . . . . . . . . . . . . . . . . . }}\\ \scriptsize{\texttt{\fbox{003:} . car nearby vehicle . . . . . . . . . . . . . . . . . . . . . . . . . . . . . . . . . . . . . . . . . . . . . . . . . . . . . . . . . . . . . . . .}}\\ \scriptsize{\texttt{\fbox{004:} . into at to . . . , nearby . . . . . . . . . . . . . . . . . . . . . . . . . . . . . . . . . . . . . . . . . . . . . . . . . . . . . . . . . . . . . }}\\ \scriptsize{\texttt{\fbox{005:} . car some because because some because some because because because because some . . . . .}}}\\%
        \bottomrule%
    \end{tabularx}%
    \caption{Generated Scene Descriptions in Stage 2 Based on \textit{seen\_bdd\_00x} Images - Part 2}\label{tab:s2-seen-2}%
\end{table}%
\thispagestyle{empty}%
\restoregeometry
\newpage%
\clearpage
\newpage%

Here, VGG-16-E01 represents VGG-16 with self-attention applied after its last feature extraction layer, before feeding the features to the decoder (i.e., intra-encoder fusion); in this case, the attention mechanism uses the extracted features as query, key, and value for the multi-head attention layer, which then produces an output of the same dimension. VGG-16-E02 represents VGG-16 with cross-attention after its last feature extraction layer before feeding the features to the decoder (i.e., inter-encoder fusion); in this case, the attention mechanism takes the extracted features from VGG-16 as query and the extracted features from ResNet-50 as key and value, trying to fuse and enrich the output of the VGG-16 encoder with the output of the ResNet-50 encoder---this is called unidirectional cross-attention fusion. VGG-ResNet includes VGG-16-E02 but simultaneously adds another attention mechanism, which takes the extracted features from ResNet-50 as query and the extracted features from VGG-16 as key and value, trying to fuse and enrich the output of the ResNet-50 encoder with the output of the VGG-16 encoder; then, the outputs of both attention layers are fused via a weighted averaging operator with trainable weights, and the output of this operator serves as the input to the decoder---this is called bidirectional cross-attention fusion.%

On the other hand, RT-DETR employs a multi-scale feature extraction technique in its hybrid encoder to process images at various resolutions, enabling the model to capture both fine-grained details (useful for detecting small objects) and high-level semantic information (useful for larger objects) \cite{2024_Zhao_RT-DETR}. Accordingly, RT-DETR-E01 captures fine details for detecting smaller objects, RT-DETR-E02 captures intermediate-level features, and RT-DETR-E03 captures larger objects and broader contexts.%

The reason for selecting VGG-16 for Stage 2 is its straightforward design and its status as one of the first networks employed in image captioning. The reason behind selecting RT-DETR for Stage 2 is its real-time capability.%

According to Table~\ref{tab:s2-seen-2}, the descriptions generated by different scales of features extracted from the encoder of RT-DETR do not make any sense, and the assumption that the output of the encoder of RT-DETR would lead to improvement in description prediction fails. As for the VGG-16-based models, it is evident from Table~\ref{tab:s2-seen} that the sentences are clear and make sense; therefore, Table~\ref{tab:s2-seen-eval} is provided for a more in-depth evaluation. Based on the scores provided in Table~\ref{tab:s2-seen-eval} as well as human judgment, VGG-16-E01 exhibited a better performance over other models.%

\begin{table}[!t]
    \centering
    \begin{tabular}{llll}
        \toprule%
        \textbf{id} & \makecell[l]{\textbf{VGG-16-E01}\\ \textbf{CIDEr $\mid$ SPICE}} & \makecell[l]{\textbf{VGG-16-E02}\\ \textbf{CIDEr $\mid$ SPICE}} & \makecell[l]{\textbf{VGG-ResNet}\\ \textbf{CIDEr $\mid$ SPICE}}\\%
        \midrule%
        {001} & {1.0000} $\mid$ {0.7499} & {1.0000} $\mid$ {0.7499} & {1.0000} $\mid$ {0.1818}\\%
        {002} & {1.0003} $\mid$ {0.1333} & {0.0000} $\mid$ {0.5625} & {1.0003} $\mid$ {0.1333}\\%
        {003} & {1.6809} $\mid$ {0.2083} & {1.6407} $\mid$ {0.2448} & {1.4209} $\mid$ {0.3773}\\%
        {004} & {1.0000} $\mid$ {0.0952} & {1.0000} $\mid$ {0.0952} & {1.0000} $\mid$ {0.0952}\\%
        {005} & {1.7802} $\mid$ {0.4375} & {1.1083} $\mid$ {0.1481} & {1.1083} $\mid$ {0.1481}\\%
        \bottomrule%
    \end{tabular}%
    \caption{Evaluation Scores for Generated Scene Descriptions in Stage 2 Based on \textit{seen\_bdd\_00x} Images}\label{tab:s2-seen-eval}%
\end{table}%

The purpose of comparing RT-DETR with VGG was to assess whether a hybrid CNN-ViT-based model can outperform a fundamental baseline in image captioning, such as VGG (or ResNet), while maintaining effectiveness. However, RT-DETR did not achieve the desired results in this evaluation. Given this outcome, it is now appropriate to explore alternative architectures. Consequently, MViTv2-S (small variant of MViTv2)---a ViT model without a CNN backbone---has been selected for further investigation, as it offers a unified approach to image and video classification as well as object detection.%

\subsubsection{Stage 3 - Analysis for Multimodal Fusion and Static Image Temporalization}%
In this stage, the encoder-decoder architecture will be finalized. Here, the comparison is made between VGG-16-E01 and MViTv2-S, which, contrary to previous encoders investigated thus far, requires an extra dimension in the input. This additional dimension accounts for the temporal aspect of the input and must be at least 16 frames. Since these experiments are conducted on images, each image is repeated 16 times and stacked on top of one another to create a static or motionless sequence of frames (i.e., static image temporalization).%

Furthermore, an additional modification is applied in the decoder before feeding the visual features and the embedded sequence of textual tokens to the xLSTM. This modification involves a cross-attention mechanism that utilizes the textual data as a query and the visual data as a key and value for the multi-head attention layer (i.e., multimodal fusion).%

According to Table~\ref{tab:s3-seen}, the descriptions generated by both models are entirely correct, and there are no extra or unrelated words in the predicted outputs. Additionally, Table~\ref{tab:s3-seen-eval} illustrates that both models are in close competition with one another. However, this must be considered in the evaluation that, due to the employment of only cross-entropy loss in the training process, it is expected to see one of the ten descriptive sentences provided as ground truth during training for each image. Therefore, if a model predicts one of the sentences correctly, it can be assumed that the model is working just fine, which is the case in Table~\ref{tab:s3-seen}. Moreover, the sentences ``\texttt{it is clear daytime .}'' or ``\texttt{it is clear nighttime .}'' have been seen the most by the models, especially the former, since most of the selected images were taken during daytime; it is, consequently, expected to see them more than other sentences in the predictions.%

\begin{table}[!t]
    \centering
    \begin{tabularx}{\textwidth}{ll}
        \toprule%
        \makecell[l]{\rotatebox{90}{\textbf{VGG-16-E01}}} & \makecell[X]{\scriptsize{\texttt{\fbox{001:} it is a residential area .}}\\ \scriptsize{\texttt{\fbox{002:} it is clear nighttime , it is a two - way street , one black suv is braking in front with its brake lights on , the ego lane is the leftmost lane , an intersection is ahead , and the traffic light is green at the intersection .}}\\ \scriptsize{\texttt{\fbox{003:} it is daytime , and the street is wet and slippery with snow .}}\\ \scriptsize{\texttt{\fbox{004:} it is clear daytime .}}\\ \scriptsize{\texttt{\fbox{005:} it is daytime , and it is raining .}}}\\%
        \midrule
        \makecell[l]{\rotatebox{90}{\textbf{MViTv2-S}}} & \makecell[X]{\scriptsize{\texttt{\fbox{001:} it is a residential area .}}\\ \scriptsize{\texttt{\fbox{002:} it is a two - way street .}}\\ \scriptsize{\texttt{\fbox{003:} a yellow taxi is driving in the ego lane some distance ahead .}}\\ \scriptsize{\texttt{\fbox{004:} it is a two - way street .}}\\ \scriptsize{\texttt{\fbox{005:} it is a one - way , narrow street , and the street is wet and slippery because of the rain .}}}\\%
        \bottomrule%
    \end{tabularx}%
    \caption{Generated Scene Descriptions in Stage 3 Based on \textit{seen\_bdd\_00x} Images}\label{tab:s3-seen}%
\end{table}%

\begin{table}[!t]
    \centering
    \begin{tabular}{lll}
        \toprule%
        \textbf{id} & \makecell[l]{\textbf{VGG-16-E01}\\ \textbf{CIDEr $\mid$ SPICE}} & \makecell[l]{\textbf{MViTv2-S}\\ \textbf{CIDEr $\mid$ SPICE}}\\%
        \midrule%
        {001} & {1.0000} $\mid$ {0.1250} & {1.0000} $\mid$ {0.1250}\\%
        {002} & {1.0007} $\mid$ {0.7826} & {1.0016} $\mid$ {0.1333}\\%
        {003} & {1.6851} $\mid$ {0.2448} & {1.0044} $\mid$ {0.2448}\\%
        {004} & {1.0000} $\mid$ {0.0952} & {1.0000} $\mid$ {0.0952}\\%
        {005} & {1.1083} $\mid$ {0.1481} & {1.7802} $\mid$ {0.4375}\\%
        \bottomrule%
    \end{tabular}%
    \caption{Evaluation Scores for Generated Scene Descriptions in Stage 3 Based on \textit{seen\_bdd\_00x} Images}\label{tab:s3-seen-eval}%
\end{table}%

To take this evaluation even further, Table~\ref{tab:s3-unseen} and Table~\ref{tab:s3-flickr8k} are provided to evaluate the performance of models on \textit{unseen\_bdd\_00x} and \textit{flickr8k\_00x} images presented in Table~\ref{tab:dataset-10kimages} and Table~\ref{tab:dataset-flickr8k}, respectively. Since there are no reference descriptions for these images, the evaluation is conducted only through human judgment. As mentioned earlier, the purpose of evaluation on \textit{unseen\_bdd\_00x} is to test generalization ability of the models, whereas the purpose of evaluation on \textit{flickr8k\_00x} is a step beyond generalization to see if the models can identify or detect any traffic-related entities in the \textit{flickr8k\_00x} images that they saw during training. This comparison shows that VGG-16-E01 tends to generate more unrelated text than MViTv2-S, but it mostly produces correct information about the image. Finally, due to its optimal performance and ability to avoid generating unrelated text, as well as its capability to work with videos, MViTv2-S is selected as the primary component of the encoder. At the same time, xLSTM serves as the primary component of the decoder, finalizing the encoder-decoder traffic scene understanding architecture.%

\clearpage%
\newpage%
\newgeometry{left=0.2in, right=0.2in, top=0.2in, bottom=0.2in}%
\begin{table}[!t]
    \centering
    \begin{tabularx}{\textwidth}{ll}
        \toprule%
        \makecell[l]{\rotatebox{90}{\textbf{VGG-16-E01}}} & \makecell[X]{\scriptsize{\texttt{\fbox{001:} it is clear daytime , it is a two - way street , and the traffic light is green .}}\\ \scriptsize{\texttt{\fbox{002:} it is clear daytime , it is a residential area , many vehicles are parked on both sides of the street , a [ school zone ] sign is on the right side of the street , and a [ school zone ] sign is on the right side of the street , and the street is wet and slippery because of the snow .}}\\ \scriptsize{\texttt{\fbox{003:} it is clear daytime , it is a two - way street separated by double solid yellow lines , a car is driving in the ego lane ahead , several vehicles are parked on both sides of the street , a car is driving in the ego lane ahead , several cars are parked on both sides of the street , a car is driving in the ego lane ahead , several cars are parked on both sides of the street , a car is driving in the ego lane ahead , several cars are parked on both sides of the street , a car is driving in the ego lane ahead , several cars are parked on both sides of the street , a car is driving in the ego lane ahead , several cars are parked on both sides of the street , a car is driving in the ego lane ahead , several cars are parked on both sides of the street , a car is driving in the ego lane ahead , several cars are parked on both sides of the street , and there are solid double yellow lines .}}\\ \scriptsize{\texttt{\fbox{004:}  it is clear daytime , it is a two - way street separated by double solid yellow lines , and a white car is driving in the right lane nearby .}}\\ \scriptsize{\texttt{\fbox{005:} it is clear nighttime , it is clear nighttime , it is in front , it is clear nighttime , it is clear nighttime , it is clear nighttime , it is clear nighttime , it is in front , it is clear nighttime , it is clear nighttime , it is clear nighttime , it is clear nighttime , it is in front , it is clear nighttime , it is clear nighttime , it is clear nighttime , it is clear nighttime , it is an intersection , the road are braking with their brake lights on , 1 vehicle with its taillights on is braking with their brake lights on an along the car is driving in front , 2 vehicles are driving in the left lane , 1 white vehicle is braking with its brake lights on in front , 1 white vehicle is braking with their brake lights on in front , 1 white vehicle is braking with their brake lights on in front , 1 white vehicle is braking with their brake lights on in front , 1 white vehicle is braking with their brake lights on in front , 1 white vehicle is braking with their brake lights on in front , 1 white vehicle is braking with their brake lights on in front , 1 white vehicle is braking with their brake lights on in front ,}}}\\%
        \midrule
        \makecell[l]{\rotatebox{90}{\textbf{MViTv2-S}}} & \makecell[X]{\scriptsize{\texttt{\fbox{001:} it is a one - way street with 2 lanes .}}\\ \scriptsize{\texttt{\fbox{002:} it is clear daytime , the street is wet and slippery with snow , and it is a residential area .}}\\ \scriptsize{\texttt{\fbox{003:} it is a residential area , and the ego lane is clear .}}\\ \scriptsize{\texttt{\fbox{004:} it is clear daytime , it is a residential area , and a yellow taxi is driving in the opposite direction in the left lane .}}\\ \scriptsize{\texttt{\fbox{005:} it is clear nighttime , it is a two - way street , and there is traffic congestion .}}}\\%
        \bottomrule%
    \end{tabularx}%
    \caption{Generated Scene Descriptions in Stage 3 Based on \textit{unseen\_bdd\_00x} Images}\label{tab:s3-unseen}%
\end{table}%
\thispagestyle{empty}%
\restoregeometry
\newpage%
\clearpage
\newpage%

\begin{table}[!t]
    \centering
    \begin{tabularx}{\textwidth}{ll}
        \toprule%
        \makecell[l]{\rotatebox{90}{\textbf{VGG-16-E01}}} & \makecell[X]{\scriptsize{\texttt{\fbox{001:} it is clear daytime , it is an intersection , the traffic light is green , a [ <UNK> is sign is on the right side of the street , 1 person is standing in front , and 1 silver car is braking with its brake lights on .}}\\ \scriptsize{\texttt{\fbox{002:} it is clear nighttime , it is a car are driving ahead in a car in a car are driving in the ego lane straight ahead , some cars are driving in the ego lane straight ahead , some cars are driving in the ego lane straight ahead , some cars are driving in the ego lane straight ahead , some cars are driving in the ego lane straight ahead , some cars are driving in the ego lane straight ahead , some cars are driving in the ego lane straight ahead , some cars are driving in the ego lane straight ahead , some cars are driving in the ego lane straight ahead , some cars are driving in the ego lane straight ahead , some cars are driving in the ego lane straight ahead , some cars are driving in the ego lane straight ahead , and the cones are on the right side of the street .}}\\ \scriptsize{\texttt{\fbox{003:} it is clear daytime , it is a narrow , one - intersection is on the right side of the street , and a black car is braking with its brake lights on in the ego lane in front .}}\\ \scriptsize{\texttt{\fbox{004:} it is clear daytime , and many vehicles are driving in the left lanes .}}\\ \scriptsize{\texttt{\fbox{005:} it is clear daytime , it is an intersection , the ego lane is the leftmost lane , a [ <UNK> ] sign is in front , some vehicles are parked on the right side of the road , a black car is driving in the ego lane at a short distance away , and 1 car is driving in the ego lane at a short distance away .}}}\\%
        \midrule
        \makecell[l]{\rotatebox{90}{\textbf{MViTv2-S}}} & \makecell[X]{\scriptsize{\texttt{\fbox{001:} it is clear daytime , it is an intersection , and 1 person is crossing the street using the crosswalk .}}\\ \scriptsize{\texttt{\fbox{002:} it is clear daytime , it is a residential area , 1 black vehicle is stopped with its brake lights on in front at a short distance away , some vehicles are driving in the right lane , 1 pedestrian is crossing the street ahead , some vehicles are driving with their brake lights on .}}\\ \scriptsize{\texttt{\fbox{003:} it is clear daytime , it is a one - way street , one pedestrian is crossing the crosswalk from left to right , and one pedestrian is crossing the crosswalk from left to right .}}\\ \scriptsize{\texttt{\fbox{004:} it is clear daytime , and many vehicles are driving on the road .}}\\ \scriptsize{\texttt{\fbox{005:} it is clear daytime , and it is clear daytime .}}}\\%
        \bottomrule%
    \end{tabularx}%
    \caption{Generated Scene Descriptions in Stage 3 Based on \textit{flickr8k\_00x} Images}\label{tab:s3-flickr8k}%
\end{table}%

\subsection{Limitations and Considerations}%
Several factors should be considered in interpreting the findings of this research, as outlined below:%
\begin{itemize}
    \item At the time of manuscript submission, the annotated dataset comprises 600 images. An additional batch of 400 images is currently being annotated according to the same criteria and will be incorporated into the dataset, resulting in a total of 1,000 annotated images. This expanded dataset will be available for future research that builds upon the present study.%
    \item This study only employed cross-entropy loss for model training. However, reinforcement learning (RL)-based fine-tuning is required to further enhance performance. Future work will address this limitation by incorporating the expanded dataset and implementing a novel RL-based fine-tuning approach.%
    \item All models were trained on the Narval cluster of the Digital Research Alliance of Canada (the Alliance), utilizing a single NVIDIA A100-SXM4-40GB GPU. As described in Subsubsection~\ref{subsubsec:TC}, the encoder parameters were frozen for the initial 10 epochs to facilitate proper decoder initialization. During the training of MViTv2-S-xLSTM, unfreezing the MViTv2-S weights at the start of epoch 11 resulted in a crash due to GPU memory exhaustion. Consequently, throughout the entire training process, only the Intermediate Bridge and decoder layers were updated, while the MViTv2-S weights remained frozen. Notably, MViTv2-S was initialized with pre-trained weights from its training on the Kinetics-400 dataset for video classification, where it achieved a state-of-the-art accuracy of 86.1\%.%
\end{itemize}

\section{Conclusion and Future Directions}%
\label{sec:s05}%
Accurate understanding of the traffic scene is fundamental to safe navigation and effective decision-making in autonomous driving. This study addressed this challenge by proposing an image-to-text approach for generating natural language descriptions of static traffic scenes. Specifically, the present study developed a model that combines MViTv2-S and xLSTM networks, augmented with hybrid attention fusion and static image temporalization, to capture critical scene features and produce contextually meaningful descriptions. Given the absence of suitable publicly available datasets for this task, this study developed a dedicated dataset, derived from the BDD100K dataset, tailored to encompass all essential aspects of driving scenes. This dataset was developed using carefully designed guidelines that prioritize safety and reliability. The annotation process was outsourced to individuals with driving experience to ensure contextual accuracy in the reference descriptions. Additionally, a comprehensive analysis of image captioning evaluation metrics was conducted to identify the most suitable metrics (i.e., CIDEr and SPICE) for evaluating traffic scene descriptions. These insights facilitate the integration of robust metrics into future training pipelines, supporting more reliable model assessments.%

Overall, the proposed framework demonstrates the effectiveness of advanced attention-based architectures for traffic scene understanding and establishes a strong foundation for further research in natural language-based scene interpretation for autonomous vehicles. Therefore, future work will focus on expanding the developed dataset to enhance model generalizability, incorporating a novel reinforcement learning-based approach for fine-tuning the model, and extending the framework to video (dynamic) data by leveraging spatiotemporal feature extraction. These directions aim to further improve the robustness and applicability of natural language scene descriptions in dynamic and complex driving environments.

\section{Declaration of Generative AI and AI-Assisted Technologies in the Writing Process}%
The authors utilized GPT-4.1 to assist in refining the grammar and lexical quality of this manuscript during the preparation and writing process. Following the application of this tool/service, the authors thoroughly reviewed and edited the manuscript to ensure accuracy and clarity, and they accept full responsibility for the final content of the publication.%

\appendix%
\section{Guidelines for Data Annotators}%
\label{app-GDA}%
\begin{itemize}
    \item \textbf{Guideline~001:}~Clearly describe all critical elements of the traffic scene relevant to safe navigation and decision-making of an autonomous vehicle, including vehicles, pedestrians, cyclists, road conditions, lane markings, traffic signs, signals, obstacles, and environmental conditions.%
    \item \textbf{Guideline 002:} Sparingly use ``\textit{There is}'' or ``\textit{There are}.''%
    \item \textbf{Guideline 003:} Do not describe irrelevant or minor visual details that do not impact driving safety or decision-making.%
    \item \textbf{Guideline 004:} Only describe what is explicitly visible in the current image. Do not speculate about events that may have occurred before or might happen afterward.%
    \item \textbf{Guideline 005:} Do not include descriptions of what people might say or conversations they might have.%
    \item \textbf{Guideline 006:} Do not assign proper names to any individual present in the scene, and do not assume genders. Use ``\textit{people},'' ``\textit{person},'' ``\textit{pedestrian},'' ``\textit{individual},'' ``\textit{adult},'' and ``\textit{child}'' instead.%
    \item \textbf{Guideline 007:} Write 10 descriptive sentences/descriptions per image. The sentences must be short and long.%
    \item \textbf{Guideline 008:} At least one sentence must explicitly state weather conditions (clear, rainy, snowy, foggy) and lighting conditions (daytime or nighttime). For example, ``\textit{it is clear daytime.}'' or ``\textit{it is clear nighttime.}'' or ``\textit{it is daytime/nighttime, and it is snowing/raining.}''%
    \item \textbf{Guideline 009:} ONE DESCRIPTION TO RULE THEM ALL - At least one sentence must be complex and lengthy, containing all critical elements described in previous sentences. It can be a combination of other sentences.%
    \item \textbf{Guideline 010:} Short sentences must focus on one or two critical aspects of the traffic scene relevant to driving decisions and potential hazards.%
    \item \textbf{Guideline 011:} Clearly specify relative positions (left/right/front/rear/ahead/behind) and relative distances (near/far/immediate vicinity) of traffic entities with respect to the ego vehicle, and explicitly highlight potential dangers or risks ahead.%
    \item \textbf{Guideline 012:} Explicitly mention colors and numbers of relevant traffic entities when they significantly aid in distinguishing entities or understanding the scene.%
    \item \textbf{Guideline 013:} Do not use single (`') or double (``'') quotation marks in the sentences.%
    \item \textbf{Guideline 014:} Avoid contractions entirely (e.g., use ``\textit{it is}'' instead of ``\textit{it's}'' or ``\textit{they are}'' instead of ``\textit{they're}'').%
    \item \textbf{Guideline 015:} Use domain-specific vocabulary provided in the examples and the document.%
    \item \textbf{Guideline 016:} End sentences with period (``.'') (e.g., ``\textit{The traffic light is green.}'').%
    \item \textbf{Guideline 017:} Use comma (``,'') to separate three or more items, placing ``\textit{, and}'' before the last item (e.g., ``\textit{a car, a pedestrian, and a cyclist}'').%
    \item \textbf{Guideline 018:} Ensure each image has exactly 10 descriptive sentences, including both short and detailed ones. If you are short on details to mention, paraphrase the existing sentences or combine them to fulfill this requirement.%
    \item \textbf{Guideline 019:} Explicitly mention all visible traffic signs and signals along with their relative positions with respect to the ego vehicle. Clearly indicate traffic sign names by enclosing them within square brackets (e.g., ``\textit{[pedestrian crossing] sign}'').%
    \item \textbf{Guideline 020:} Represent numbers using digits (e.g., ``\textit{2 pedestrians}'') rather than words. However, you can interchangeably use ``\textit{one},'' ``\textit{a},'' ``\textit{an},'' or ``\textit{1}.''%
    \item \textbf{Guideline 021:} Do not use Generative AI to generate any part of the sentences. The sentences must be completely human-generated.%
    \item \textbf{Guideline 022:} Consistently use the American English terminology for traffic-related terms (e.g., ``\textit{truck}'' instead of ``\textit{lorry},'' ``\textit{crosswalk}'' instead of ``\textit{pedestrian crossing},'' ``\textit{railroad crossing}'' instead of ``\textit{railway crossing},'' or ``\textit{taillights}'' instead of ``\textit{tail lights}''). When uncertain, verify terms using the Merriam-Webster dictionary or simply Google.%
    \item \textbf{Guideline 023:} For images that are part of a sequence, reuse identical sentences for consecutive images if no significant changes or new important details appear between frames.%
    \item \textbf{Guideline 024:} Clearly indicate lane configurations (number of lanes, merging lanes, turning lanes) and lane-specific details such as lane closures or construction zones if visible.%
    \item \textbf{Guideline 025:} Clearly describe road markings, especially if they are hardly visible, such as lane lines (solid/dashed), stop lines, crosswalk markings, arrows indicating permitted direction of travel, bike lanes, or special road markings relevant to autonomous driving decisions.%
    \item \textbf{Guideline 026:} Explicitly describe observable behaviors of other vehicles or road users that could impact safety (e.g., sudden braking of vehicles ahead, pedestrians stepping into the roadway).%
    \item \textbf{Guideline 027:} Explicitly mention road surface conditions when relevant to safety or decision-making (e.g., wet roads, icy patches, potholes).%
    \item \textbf{Guideline 028:} Clearly describe any obstructions or visibility limitations present in the scene (e.g., parked vehicles blocking the view at intersections, vegetation obstructing signs).%
    \item \textbf{Guideline 029:} Explicitly mention the presence and position of emergency vehicles with activated lights/sirens or other special situations requiring immediate attention.%
    \item \textbf{Guideline 030:} Explicitly describe visible turn signals/brake lights/reverse lights from other vehicles when relevant to predicting their actions.%
    \item \textbf{Guideline 031:} Clearly highlight pedestrian/cyclist behaviors that may indicate potential risk (e.g., jaywalking pedestrians or cyclists preparing to cross).%
    \item \textbf{Guideline 032:} Maintain consistency in terminology and units across all descriptions; use standard American English terms consistently as per Guideline 022.%
    \item \textbf{Guideline 033:} Maintain consistency in terminology by using ``\textit{vehicle}'' interchangeably with ``\textit{car},'' ``\textit{truck},'' ``\textit{bus},'' ``\textit{motorcycle},'' or ``\textit{bicycle},'' especially when describing a mix of different types.%
    \item \textbf{Guideline 034:} All descriptions must be written clearly from the perspective of the ego vehicle driver position; avoid descriptions from external viewpoints or aerial perspectives.%
\end{itemize}

\bibliographystyle{elsarticle-num}%
\bibliography{references}%
\end{document}